%% file: bare_jrnl.tex
\documentclass[10pt,twocolumn,twoside]{IEEEtran}
\usepackage{cite}

\ifCLASSINFOpdf
   \usepackage[pdftex]{graphicx}
   \graphicspath{images/}
   \DeclareGraphicsExtensions{.pdf,.jpeg,.png}
\else
\fi
\usepackage[font=footnotesize,subrefformat=parens,labelformat=parens]{subfig}

\usepackage{dsfont}
\usepackage{mathtools,amsmath}
\usepackage{mathrsfs}
\usepackage{amssymb}
\usepackage{bm}
\usepackage{xcolor}

\newcommand{\new}[1]{{\color{black}{#1}}}
\DeclareMathOperator*{\argmin}{argmin}
\usepackage{algorithm}
\usepackage{algpseudocode}
\usepackage{siunitx}

\begin{document}
%
\title{Connectivity Maintenance for Multi-Robot Systems Under Motion and Sensing Uncertainties Using Distributed ADMM-based Trajectory Planning}
%
%

\author{Akshay~Shetty, Derek Knowles
        and~Grace~Xingxin~Gao
\thanks{Akshay Shetty is with the Department of Aerospace Engineering, University of Illinois at Urbana-Champaign, Champaign, IL, 61801, USA. e-mail: ashetty2@illinois.edu.}
\thanks{Derek Knowles is with the Department of Mechanical Engineering, Stanford University, CA, 94305, USA. e-mail: dcknowles@stanford.edu.}
\thanks{Grace Xingxin Gao is with the Department of Aeronautics and Astronautics, Stanford University, CA, 94305, USA. e-mail: gracegao@stanford.edu.}
\thanks{Manuscript received month xx, 20xx; revised month xx, 20xx.}}

\maketitle

\begin{abstract}
\input{abstract}
\end{abstract}

\vspace{-0.15cm}

%
\IEEEpeerreviewmaketitle

\vspace{-0.35cm}
\section{Introduction}
\input{introduction}

\vspace{-0.2cm}
\section{Related Work: Connectivity Maintenance}
\input{related_work}

\vspace{-0.2cm}
\section{Problem Formulation}
\input{problem_formulation}

\vspace{-0.3cm}
\section{Weighted Undirected Graph For Uncertain Robot Positions}
\input{connectivity}

\vspace{-0.2cm}
\section{Trajectory Planning Algorithm}
\input{trajectory_planning}

\vspace{-0.3cm}
\section{Simulations}
\input{simulations}

\vspace{-0.4cm}
\section{Conclusions}
\input{conclusions}
\ifCLASSOPTIONcaptionsoff
  \newpage
\fi



%

\bibliographystyle{IEEEtran}
\bibliography{references}
%
\vspace{-1.5cm}
\begin{IEEEbiography}[{\includegraphics[width=1in,height=1.25in,clip,keepaspectratio]{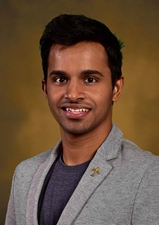}}]{Akshay Shetty}
is a postdoctoral researcher in the Department of Aerospace Engineering at Stanford University. He received his Ph.D. degree in Aerospace Engineering from University of Illinois at Urbana-Champaign in 2021. His research interests include safe trajectory planning and control for autonomous vehicles.
\end{IEEEbiography}
\vspace{-1.5cm}
\begin{IEEEbiography}[{\includegraphics[width=1in,height=1.25in,clip,keepaspectratio]{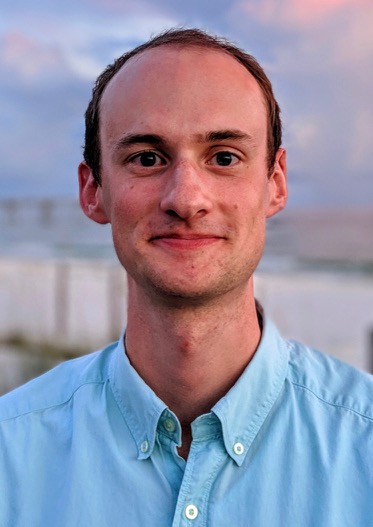}}]{Derek Knowles}
recieved the B.S. degree in mechanical engineering from Brigham Young University in 2019. He is currently pursuing the Ph.D. degree in mechanical engineering from Stanford University.

His research interests include autonomous robotic navigation, robust perception, and safe control.
\end{IEEEbiography}
\vspace{-1.5cm}
\begin{IEEEbiography}[{\includegraphics[width=1in,height=1.25in,clip,keepaspectratio]{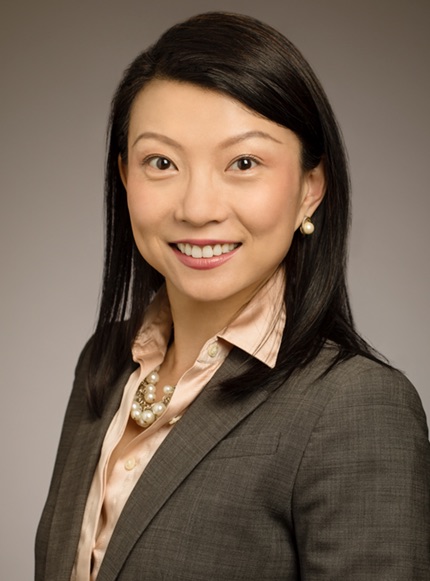}}]{Grace Xingxin Gao}
is an assistant professor in the Department of Aeronautics and Astronautics at Stanford University. Before joining Stanford University, she was an assistant professor at University of Illinois at Urbana-Champaign. She obtained her Ph.D. degree at Stanford University. Her research is on robust and secure positioning, navigation and timing with applications to manned and unmanned aerial vehicles, robotics and power systems.
\end{IEEEbiography}




\end{document}

%% file: abstract.tex
Inter-robot communication enables multi-robot systems to coordinate and execute complex missions efficiently. Thus, maintaining connectivity of the communication network between robots is essential for many multi-robot systems. In this paper, we present a trajectory planner for connectivity maintenance of a multi-robot system. We first define a weighted undirected graph to represent the connectivity of the system. Unlike previous connectivity maintenance works, we explicitly account for robot motion and sensing uncertainties while formulating the graph edge weights. These uncertainties result in uncertain robot positions which directly affect the connectivity of the system. Next, the algebraic connectivity of the weighted undirected graph is maintained above a specified lower limit using a trajectory planner based on a distributed alternating direction method of multipliers (ADMM) framework. Here we derive an approximation for the Hessian matrices required within the ADMM optimization step to reduce the computational load. Finally, simulation results are presented to statistically validate the connectivity maintenance of our trajectory planner.

%% file: introduction.tex
\label{sec:introduction}

There has been growing interest in multi-robot systems for exploration, target tracking, formation control, and cooperative manipulation~\cite{rizk2019cooperative}.
Multi-robot systems typically depend on inter-robot communication which enables them to execute complex missions efficiently. Inter-robot communication also adds resilience to malicious attacks~\cite{alanwar2019distributed} and single robot failures~\cite{park2018robust}. Thus, maintaining connectivity between robots is often a requirement for multi-robot systems.

The topic of connectivity maintenance for multi-robot systems has been widely addressed in literature. The general approach is to synthesize control inputs for each robot in the system such that either local connectivity or global connectivity of the system is maintained~\cite{khateri2019comparison}. Local connectivity maintenance (LCM) methods focus on keeping the initial topology of connections within the multi-robot system. Thus, if two robots are initially connected, the synthesized control inputs for these robots maintain their connection throughout the mission. LCM methods typically consist of relatively simple computations since the control inputs for each robot depend only on local information of the robots to which it is connected. However, the freedom of motion for each robot is restricted since initial connections between robots are not allowed to be broken. Global connectivity maintenance (GCM) methods on the other hand allow individual connections to break as long as there exists a (potentially multi-hop) communication path between two robots in the system~\cite{de2006decentralized,yang2010decentralized,Sabattini2013,Sabattini2013a,robuffo2013passivity,gasparri2017bounded,capelli2020connectivity,Bhattacharya2010}. Thus, each robot is afforded a greater freedom of motion compared to LCM methods.

A limitation of the previous connectivity maintenance works is that they do not explicitly account for robot motion and sensing uncertainties in their theoretical formulation. Presence of motion uncertainty causes a robot to deviate from the trajectory desired by the synthesized control inputs. Additionally, sensing uncertainties result in a robot obtaining a noisy estimate of its own position and its neighbors' positions. These uncertainties, which are inherent in practical robots~\cite{thrun2005probabilistic}, consequently result in uncertainty in the robot positions and directly affects the system connectivity. Thus, it is important to explicitly account for robot motion and sensing uncertainties while designing connectivity maintenance methods.

Additionally, the majority of previous connectivity maintenance works use a simplified single integrator model to represent the robot motion~\cite{de2006decentralized,yang2010decentralized,Sabattini2013,Sabattini2013a,robuffo2013passivity,gasparri2017bounded,capelli2020connectivity}. This motion model assumes that the robot can instantaneously change its direction of motion and move towards a desired position for connectivity maintenance. Thus, these works derive control inputs in a myopic fashion, i.e., only for the current time instant. However, for most practical robots, such as unmanned aerial vehicles (UAVs), the direction of motion is not instantaneously changeable. The robot trajectory depends on additional quantities (such as previous velocities~\cite{mclain2014implementing}) and hence the robot might not be able to change its direction of motion instantaneously. Thus, for connectivity maintenance, it is important to derive control inputs in a non-myopic fashion by considering the trajectory of the robot over multiple future time instants. 

In this paper, we present a trajectory planning algorithm that maintains the global connectivity of a multi-robot system while accounting for motion and sensing uncertainties. The main contributions of this paper are listed as follows:
\begin{enumerate}
    \item We define a weighted undirected graph to represent a multi-robot system with uncertain robot positions. Here the uncertainty in the robot positions are obtained from the robot motion and sensing uncertainties. We show that the algebraic connectivity (explained in Section~\ref{subsec:connectivity-maintenance}) of this graph is a probabilistic lower-bound for the true algebraic connectivity of the system.
    \item We use the algebraic connectivity of the proposed weighted undirected graph to design a distributed trajectory planning algorithm based on an alternating direction method of multipliers (ADMM) framework~\cite{Boyd2010}. Here we use a general linear motion model for the robots that does not necessarily assume an instantaneously changeable direction of motion. Our algorithm plans trajectories for the multi-robot system in a non-myopic fashion while maintaining the algebraic connectivity of the proposed graph above a specified lower limit.
    \item We analyze the computational load of the optimization step within the ADMM framework. An approximation is derived for the required Hessian matrices which significantly reduces the computational load.
    \item We simulate multi-UAV missions to statistically validate the global connectivity maintenance of our trajectory planner. We show the convergence of our algorithm and evaluate its performance under \new{real-}time constraints.
\end{enumerate}
The remainder of the paper is organized as follows. We begin by discussing related connectivity maintenance works in Section~\ref{sec:related-work}. 
In Section~\ref{sec:problem-formulation}, we formulate the connectivity maintenance and trajectory planning problems for this paper. Section~\ref{sec:connectivity-metric} defines the weighted undirected graph proposed for uncertain robot positions, which we use in our trajectory planning algorithm detailed in Section~\ref{sec:trajectory-planning}. We present our simulation results in Section~\ref{sec:simulations} and conclude in Section~\ref{sec:conclusions}.

%% file: related_work.tex
\label{sec:related-work}


\label{subsec:global-connectivity}
GCM is widely addressed in literature \new{and common methods are based on branching trees, \textit{k}-connectivity, or algebraic connectivity. In the branching tree method, multiple robots are initially clumped together, one robot moves away from the group to the edge of the connectivity range, and then the preceding robots follow that pattern extending the communication branch to reach a desired goal~\cite{Luo,Scherer}. The main drawback of the branching tree method is that only a small subset of the robots can move at a given time which limits the amount of area that can be covered by the network. \textit{k}-connectivity GCM methods design controllers that maintain a set level of \textit{k}-connectivity at all times \cite{luo2019minimum}. A robot network is \textit{k}-connected if the network remains connected if fewer than \textit{k} robots are removed from the network.


Finally, a third group of GCM methods} represent the multi-robot system as a weighted undirected graph and use its algebraic connectivity as an indicator of the system connectivity. The algebraic connectivity is defined as the second smallest eigenvalue of the graph Laplacian matrix, as discussed later in Section~\ref{subsec:connectivity-maintenance}. In~\cite{yang2010decentralized}, the authors present a decentralized power iteration algorithm for each robot to estimate the algebraic connectivity. This estimate is then used to design a decentralized gradient-based controller for GCM. \cite{Sabattini2013} builds on the method in~\cite{yang2010decentralized} by defining a decentralized estimation procedure for algebraic connectivity that is formally guaranteed to be stable. Given the estimation error boundedness, they prove that the proposed control law guarantees GCM if the control parameters are chosen appropriately. Further, in~\cite{Sabattini2013a} the authors extend their previous work of~\cite{Sabattini2013} by accounting for an additional (bounded) control input for each robot. \cite{robuffo2013passivity} extends on~\cite{Sabattini2013} by explicitly accounting for additional inter-robot constraints such as a desired relative distance and collision avoidance. In~\cite{gasparri2017bounded}, the authors design a GCM method to account for robots with bounded control inputs. They present a theoretical analysis to evaluate the robustness of the controller to bounded errors in estimate of the algebraic connectivity. However, the estimation error bound is heuristically obtained without explicitly accounting for sources of uncertainty such as robot motion and sensing uncertainties.

Another common approach for GCM is to design optimization-based methods without estimating the value of the algebraic connectivity itself. In~\cite{de2006decentralized}, the authors find optimal positions to maximize the algebraic connectivity and then derive control inputs for a multi-robot system using a decentralized potential field-based method. In~\cite{Bhattacharya2010}, the authors present a differential game-theoretic formulation for maximizing the algebraic connectivity in the presence of a malicious jammer. \cite{capelli2020connectivity} uses control barrier functions to integrate a GCM requirement with an additional control input for each robot. 

While many of these methods \new{provide GCM guarantees~\cite{luo2019minimum,yang2010decentralized,Sabattini2013,Sabattini2013a,robuffo2013passivity,gasparri2017bounded,de2006decentralized,Bhattacharya2010,capelli2020connectivity}}, they do not explicitly account for robot motion and sensing uncertainties and a majority of them assume a simplified single integrator robot motion model. Thus, in their simulation/experimental setups they make simplifications; for instance, assuming perfect sensing information such as perfect localization measurements and/or using slow-moving robots that can be reasonably modeled as single integrator systems. However, practical robots are typically represented by higher-fidelity motion models and use state estimation filters to estimate their positions under motion and sensing uncertainties~\cite{thrun2005probabilistic}. In this paper, we primarily address these limitations in previous methods. In the remainder of the paper, we simply refer to global connectivity as connectivity.

%% file: problem_formulation.tex
\label{sec:problem-formulation}

\subsection{Robot description}
\label{subsec:ekf-belief-dynamics}

For each robot $i$ in a multi-robot system with $N$ robots, we consider linear discrete-time motion and sensing models:
\vspace{-0.1cm}
\begin{align}
    \label{eqn:motion-model}
    \mathbf{x}_{i,t} &= A_{i,t-1} \mathbf{x}_{i,t-1} + B_{i,t-1} \mathbf{u}_{i,t-1} + \mathbf{w}_{i,t}, \\
    \label{eqn:measurement-model}
    \mathbf{z}_{i,t} &= C_{i,t} \mathbf{x}_{i,t} + \mathbf{v}_{i,t},
\end{align}
where $t$ is the time instant, $\mathbf{x}_{i,t}$ is the state vector, $\mathbf{u}_{i,t}$ is the input vector, $\mathbf{z}_{i,t}$ is the sensed measurement vector, $A_{i,t}$ is state transition matrix, $B_{i,t}$ is the control-input matrix, $C_{i,t}$ is the system measurement matrix, \new{$\mathbf{w}_{i,t}~\sim~\mathcal{N}[\mathbf{0},Q_{i,t}]$ is the motion model error and $\mathbf{v}_{i,t}~\sim~\mathcal{N}[\mathbf{0},R_{i,t}]$ is the sensing model error. Note that throughout the paper we use bold font to represent vectors, and we use the notation $\mathcal{N}[\bm{\mu},\Gamma]$ to represent a Gaussian-distributed vector with mean $\bm{\mu}$ and covariance $\Gamma$.}


We assume that each robot implements a Kalman Filter (KF)~\cite{thrun2005probabilistic} on board to obtain an estimate of its state $\hat{\mathbf{x}}_{i}$. The prediction step of the KF is performed as:
\vspace{-0.1cm}
\begin{align}
    \label{eqn:ekf-pred-1}
    \bar{\mathbf{x}}_{i,t} &= A_{i,t-1} \hat{\mathbf{x}}_{i,t-1} + B_{i,t-1} \mathbf{u}_{i,t-1}, \\
    \label{eqn:ekf-pred-2}
    \bar{P}_{i,t} &= A_{i,t-1} P_{i,t-1} A_{i,t-1}^\top + Q_{i,t},
\end{align}
where $P_{i,t}$ is the state estimation covariance matrix such that $\mathbf{x}_{i,t} \sim \mathcal{N}[\hat{\mathbf{x}}_{i,t}, P_{i,t}]$. The KF correction step is performed as:
\vspace{-0.1cm}
\begin{align}
    \label{eqn:ekf-update-1}
    L_{i,t} &= \bar{P}_{i,t} C_{i,t}^\top (C_{i,t} \bar{P}_{i,t} C_{i,t}^\top + {R}_{i,t})^{-1}, \\
    \label{eqn:ekf-update-2}
    \hat{\mathbf{x}}_{i,t} &= \bar{\mathbf{x}}_{i,t} + L_{i,t}(\mathbf{z}_{i,t} - C_{i,t} \bar{\mathbf{x}}_{i,t} ), \\
    \label{eqn:ekf-update-3}
    P_{i,t} &= \bar{P}_{i,t} - L_{i,t} C_{i,t} \bar{P}_{i,t},
\end{align}
where $L_i$ is the Kalman gain. Here the second term in~(\ref{eqn:ekf-update-2}) is referred to as the \textit{innovation} term and is distributed according to $\mathcal{N}[ \mathbf{0}, L_{i,t} C_{i,t} \bar{P}_{i,t} ]$.

Thus, each robot can be represented in the belief space with the belief vector defined as~\cite{van2012motion}: 
\vspace{-0.1cm}
\begin{equation}
    \label{eqn:belief-vec}
    \mathbf{b}_{i,t} = \begin{bmatrix} \hat{\mathbf{x}}_{i,t} \\ \textbf{vec}[P_{i,t}] \end{bmatrix},
\end{equation}
where $\textbf{vec}[P_{i,t}]$ denotes a column vector containing the elements of \new{the upper triangle portion of $P_{i,t}$ (element in first column, appended by elements in second column, and so on)}. Furthermore, the belief dynamics for the robot can be summarized as~\cite{van2012motion}:
\vspace{-0.1cm}
\begin{equation}
    \label{eqn:belief-dynamics}
    \mathbf{b}_{i,t+1} = \mathbf{g}_i[\mathbf{b}_{i,t}, \mathbf{u}_{i,t}] + M_i[\mathbf{b}_{i,t}, \mathbf{u}_{i,t}]\mathbf{m}_{i,t},
\end{equation}
\vspace{-0.05cm}
where:
\vspace{-0.5cm}
\begin{equation}
\nonumber
\begin{gathered}
    \mathbf{g}_i[\mathbf{b}_{i,t}, \mathbf{u}_{i,t}] = \begin{bmatrix} \bar{\mathbf{x}}_{i,t} \\ \textbf{vec}[\bar{P}_{i,t} - L_{i,t} C_{i,t} \bar{P}_{i,t}] \end{bmatrix} , \\
    M_i[\mathbf{b}_{i,t}, \mathbf{u}_{i,t}] = \begin{bmatrix} \sqrt{L_{i,t} C_{i,t} \bar{P}_{i,t}}\\ 0 \end{bmatrix}, \\
    \mathbf{m}_{i,t} \sim \mathcal{N}[\mathbf{0}, \mathcal{I}],
\end{gathered}
\end{equation}
where $\mathcal{I}$ represents an identity matrix.

\new{Given that we assume linear models in~(\ref{eqn:motion-model})-(\ref{eqn:measurement-model}), the KF exactly represents the uncertainty in the true state as $\mathbf{x}_{i,t} \sim \mathcal{N}[\hat{\mathbf{x}}_{i,t}, P_{i,t}]$~\cite{thrun2005probabilistic}, and consequently the belief dynamics in~(\ref{eqn:belief-dynamics}) exactly captures the state uncertainty. While the belief dynamics can be derived for nonlinear models with an Extended Kalman Filter (EKF) (as done in~\cite{van2012motion}), the EKF only provides an approximation of the state uncertainty. Thus, designing the trajectory planner based on an approximation of the state uncertainty could lead to undesirable loss of connectivity. While the linear model in~(\ref{eqn:motion-model}) is more restrictive (as opposed to a nonlinear model), it represents the motion of robotic systems more realistically~\cite{hou2015distributed,park2019distributed} compared to a single-integrator motion model assumed in a majority of related work (see Section~\ref{sec:related-work}). The linear sensing model in~(\ref{eqn:measurement-model}) is commonly used to represent measurements from on-board sensors, such as localization measurements from Global Navigation Satellite System (GNSS) receivers or from cameras. For our simulations in Section~\ref{sec:simulations}, we use a double-integrator motion model (state vector contains robot position and velocity; input vector contains accelerations) along with localization measurements.}

\vspace{-0.3cm}
\subsection{Connectivity maintenance}
\label{subsec:connectivity-maintenance}

Similar to most previous connectivity maintenance works~\cite{yang2010decentralized,Sabattini2013,Sabattini2013a,robuffo2013passivity,gasparri2017bounded}, we assume a disk communication model. Thus, two robots are considered to be connected only if the distance between them is smaller than a specified communication range $\Delta$. \new{Let the multi-robot system be represented as an undirected graph, where each node represents a robot and each edge represents the connection between two robots. The adjacency matrix of the graph at any time-step $t$ can be obtained as:}
\vspace{-0.1cm}
\begin{equation}
    \label{eqn:binary-edge-weight}
    \mathcal{A}_{ij,t} = \begin{cases} 1 & 0 \leq l_{ij,t} \leq \Delta \\
    0 & l_{ij,t} > \Delta
\end{cases},
\end{equation}
\new{where $\mathcal{A}_{ij,t}$ is the $(i,j)^{th}$ element of adjacency matrix $\mathcal{A}_t$, and $l_{ij,t}$ is the distance between the robots. The distance $l_{ij,t}$ can be computed as $l_{ij,t} = \left \| {\mathbf{p}}_{i,t} - {\mathbf{p}}_{j,t} \right \|_2$, where $\mathbf{p}_{i}$ and $\mathbf{p}_{j}$ are the true positions of the robots and $\left \| \cdot \right \|_2$ represents the L2-norm. Here we assume that the robot positions $\mathbf{p}_{i}$ are contained in the robot state vectors $\mathbf{x}_{i}$ (defined in (\ref{eqn:motion-model})), which is generally true for most mobile robot systems such as UAVs.}

\new{Given the adjacency matrix, the degree of each node can be obtained as $d_{i,t} = \sum_{j=1}^{N} \mathcal{A}_{ij,t}$. The vector of node degrees $\mathbf{d}_{t}$ is then used to define the degree matrix $\mathcal{D}_{t}$ of the graph as $\mathcal{D}_{t} = \texttt{diag}(\mathbf{d}_{t})$. Using matrices $\mathcal{A}_{t}$ and $\mathcal{D}_{t}$ the Laplacian matrix $\mathcal{L}_{t}$ of the graph is defined as $\mathcal{L}_{t} = \mathcal{D}_{t} - \mathcal{A}_{t}$~\cite{grone1990laplacian}. The second-smallest eigenvalue of the Laplacian matrix, $\lambda_2^{\mathcal{L}_t}$, is defined as the algebraic connectivity of the graph, which is a commonly used indicator for connectivity as discussed in Section~\ref{sec:related-work}. The value of $\lambda_2^{\mathcal{L}_t}$ varies from zero (if the graph is disconnected) to the number nodes in the graph (if the graph is fully connected), i.e. $0 \leq \lambda_2^{\mathcal{L}_t} \leq N$. Thus, $\lambda_2^{\mathcal{L}_t} > 0$ implies that multi-robot system is connected, i.e., there exists a (potentially multi-hop) communication path between any two robots.}

\new{Note that the value of $\lambda_2^{\mathcal{L}_t}$ depends on the robot positions $\mathbf{p}_{i}$ which are contained in the state vectors $\mathbf{x}_{i}$. Since the state vectors are stochastic in nature (as discussed in Section~\ref{subsec:ekf-belief-dynamics}), the value of $\lambda_2^{\mathcal{L}_t}$ is also stochastic.}
Thus, given a desired lower limit $\epsilon$ for the algebraic connectivity of the system, we state the following connectivity maintenance requirement for our trajectory planning algorithm:
\vspace{-0.1cm}
\begin{equation}
    \label{eqn:connectivity-maintenance}
    \Pr[ \lambda_2^{\mathcal{L}_t} > \epsilon ] \geq 1 - \delta \ \forall \ t \in [0,T],
\end{equation}
i.e., the planner should maintain $\lambda_2^{\mathcal{L}_t}$ above $\epsilon$ with a minimum probability value of \new{$(1-\delta)$} for the planning time horizon $T$. We specify the values of $\epsilon$ and $\delta$ chosen for our simulations later in Section~\ref{subsec:sim-setup}.

\vspace{-0.3cm}
\subsection{Trajectory planning}
\label{subsec:trajecory-planning}
The objective of the trajectory planner is to plan nominal trajectories for each robot such that they perform local tasks while maintaining connectivity within the multi-robot system. Here the local tasks can represent objectives such as tracking a target, minimizing the control input effort, avoiding collisions, reaching a desired position for exploration, coverage or formation control, etc. We assume that the following information is available to each robot in the system:
\begin{enumerate}
    \item \new{The number of robots in the system $N$, and} the initial beliefs of all robots, i.e., $\mathbf{b}_{i,\text{init}}$~$\forall$~$i~\in~[1,N]$. As defined in~(\ref{eqn:belief-vec}), the initial belief vector consists of the initial state estimate and the initial estimation covariance.
    
    \item The belief dynamics associated with all robots in the system as defined in~(\ref{eqn:belief-dynamics}).
    
    \item The cost functions representing the local tasks for all robots in the system, i.e., $J_{i,t}[{\mathbf{b}}_{i,t}, {\mathbf{u}}_{i,t}] \ \forall~i~\in~[1,N], \forall~t~\in~[0,T]$.
\end{enumerate}
The nominal trajectory for each robot $i$ can be represented as a series of nominal beliefs and nominal control inputs $( \check{\mathbf{b}}_{i,0}, \check{\mathbf{u}}_{i,0}, \dots , \check{\mathbf{b}}_{i,T-1}, \check{\mathbf{u}}_{i,T-1}, \check{\mathbf{b}}_{i,T} )$ \cite{van2012motion}, such that:
\vspace{-0.1cm}
\begin{equation}
    \label{eqn:nominal-dynamics}
    \check{\mathbf{b}}_{i,t+1} = \mathbf{g}_i[\check{\mathbf{b}}_{i,t}, \check{\mathbf{u}}_{i,t}]\ \forall \ t \in [0,T-1].
\end{equation}
We define a concatenated nominal input matrix $\check{U}$, consisting of nominal input vectors of all robots in the system, over the entire planning time horizon:
\vspace{-0.1cm}
\begin{equation}
\check{U} = \begin{bmatrix}
\check{\mathbf{u}}_{1,0} & \dots & \check{\mathbf{u}}_{1,T-1} \\ 
\vdots & \vdots & \vdots \\ 
\check{\mathbf{u}}_{N,0} & \dots & \check{\mathbf{u}}_{N,T-1}
\end{bmatrix}.
\end{equation}
Note that given the initial beliefs $\mathbf{b}_{i,\text{init}} \ \forall \ i \in [1,N]$, it is sufficient to represent the nominal trajectories for the multi-robot system by $\check{U}$ since the nominal beliefs for each robot can be calculated recursively using~(\ref{eqn:nominal-dynamics}). \new{Thus, in the remainder of the paper we simply refer to the concatenated nominal input matrix $\check{U}$ as the nominal trajectory for the multi-robot system.}

\new{Finally}, the overall objective of the planner is stated as:
\vspace{-0.1cm}
\begin{equation}
\begin{gathered}
    \vspace{-0.2cm}
    \check{U} = \argmin_{} \sum_{t = 0}^{T} \sum_{i = 1}^{N} J_{i,t}[\check{\mathbf{b}}_{i,t}, \check{\mathbf{u}}_{i,t}],\\
    \label{eqn:objective} \text{subject to:}  \hspace{5cm} \\
    \Pr[ \lambda_2^{\mathcal{L}_t} > \epsilon ] \geq 1-\delta \ \forall \ t \in [0,T],\\
    \check{\mathbf{b}}_{i,0} = \mathbf{b}_{i,\text{init}} \ \forall \ i \in [1,N],\\
    \check{\mathbf{b}}_{i,t+1} = \mathbf{g}_i[\check{\mathbf{b}}_{i,t}, \check{\mathbf{u}}_{i,t}] \ \forall \ i \in [1,N], \forall \ t \in [0,T-1],
\end{gathered}
\end{equation}
where the first constraint is the connectivity maintenance requirement stated in~(\ref{eqn:connectivity-maintenance}). \new{Common examples for the cost functions $J_{i,t}$ include distance to a desired position, proximity to a unsafe set of states (such as collisions), amount of required control input, etc. In order to solve the above planning problem, we first define a weighted undirected graph in Section~\ref{sec:connectivity-metric} that accounts for uncertain robot positions, and then propose a distributed ADMM-based trajectory planning algorithm in Section~\ref{sec:trajectory-planning}.} 

%% file: connectivity.tex
\label{sec:connectivity-metric}

In order to address the connectivity maintenance requirement from~(\ref{eqn:connectivity-maintenance}), we first define a weighted undirected graph that accounts for uncertain robot positions arising due to the presence of motion and sensing uncertainties. The algebraic connectivity of this graph is then used in our trajectory planning algorithm in Section~\ref{sec:trajectory-planning}. Since the graph definition is applicable for any time instant $t \in [0,T]$, for simplicity we omit the time notations in this section. \new{As mentioned in Section~\ref{subsec:connectivity-maintenance}, we assume that the robot positions $\mathbf{p}_i$ are contained in the state vectors $\mathbf{x}_i$. Thus, given that the state vector is distributed as $\mathbf{x}_{i} \sim \mathcal{N}[\hat{\mathbf{x}}_{i}, P_{i}]$ (see Section~\ref{subsec:ekf-belief-dynamics}), the robot positions can be represented as $\mathbf{p}_i \sim \mathcal{N}[\hat{\mathbf{p}}_i, \Sigma_i]$. Here $\hat{\mathbf{p}}_i$ is the estimated position contained in the estimated state $\hat{\mathbf{x}}_i$, and the covariance matrix $\Sigma_i$ is a submatrix of $P_{i}$.}


\begin{figure}[t]
  \centering
  \includegraphics[width=0.6\linewidth]{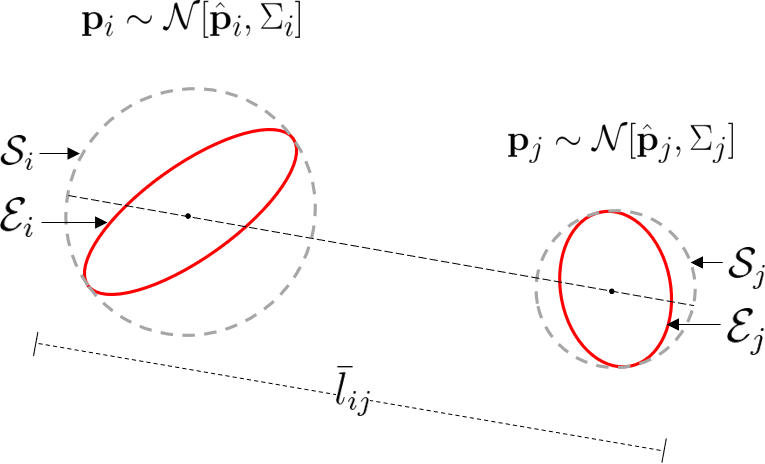}
  \caption{Distance measure $\bar{l}_{ij}$ between two robots with Gaussian-distributed positions $\mathbf{p}_{i}\sim\mathcal{N}[\hat{\mathbf{p}}_{i}, \Sigma_{i}]$ and $\mathbf{p}_{j}\sim\mathcal{N}[\hat{\mathbf{p}}_{j}, \Sigma_{j}]$. $\bar{l}_{ij}$ is the maximum distance between the boundaries of the circular regions $\mathcal{S}_i$ and $\mathcal{S}_j$ which overbound the confidence ellipsoids $\mathcal{E}_i$ and $\mathcal{E}_j$ respectively.}
  \label{fig:conservative-length}
  \vspace{-0.5cm}
\end{figure}

We begin defining our weighted graph by considering a confidence ellipse $\mathcal{E}_i$ centered at $\hat{\mathbf{p}}_{i}$ such that:
\vspace{-0.1cm}
\begin{equation}
    \label{eqn:confidence-ellipse}
    \Pr[ {\mathbf{p}}_{i} \in \mathcal{E}_i] = 1 - \delta_{\mathcal{E}}, 
\end{equation}
where $\delta_{\mathcal{E}}$ is a probability value that decides the size of the confidence ellipse. We derive the value for $\delta_{\mathcal{E}}$ used in our algorithm later in~(\ref{eqn:deltaE-value}). Let $\bar{\lambda}^{\Sigma_{i}}$ represent the largest eigenvalue of the covariance matrix $\Sigma_i$. Thus, the length of the semi-major axis of $\mathcal{E}_i$ is $s\sqrt{{\bar{\lambda}}^{\Sigma_{i}}}$, where $s$ is a scalar factor that follows a chi-square distribution~\cite{hoover1984algorithms} based on the value of $\delta_{\mathcal{E}}$. We then define a circular region $\mathcal{S}_i$ centered at $\hat{\mathbf{p}}_i$ with radius $s\sqrt{{\bar{\lambda}}^{\Sigma_{i}}}$. This circular region overbounds $\mathcal{E}_i$ and thus, contains $\mathbf{p}_i$ with a probability greater than or equal to $\delta_{\mathcal{E}}$, i.e.:
\vspace{-0.1cm}
\begin{equation}
    \label{eqn:circular-region}
    \Pr[ {\mathbf{p}}_{i} \in \mathcal{S}_i] \geq 1-\delta_{\mathcal{E}}.
\end{equation}
We then define a distance measure between the boundaries of the overbounding circular regions of two robots $i$ and $j$ as:
\vspace{-0.1cm}
\begin{equation}
    \label{eqn:conservative-dist}
    \bar{l}_{ij} = \left \| \hat{\mathbf{p}}_{i} - \hat{\mathbf{p}}_{j} \right \|_2 + s\sqrt{{\bar{\lambda}}^{\Sigma_{i}}} + s\sqrt{{\bar{\lambda}}^{\Sigma_{j}}}.
\end{equation}
Fig.~\ref{fig:conservative-length} illustrates the confidence ellipses, the overbounding circular regions and the distance measure between two robots.

\begin{figure}[t]
  \centering
  \subfloat[]{%
        \includegraphics[width=0.4\linewidth,trim={0cm 0cm 1cm 1cm},clip]{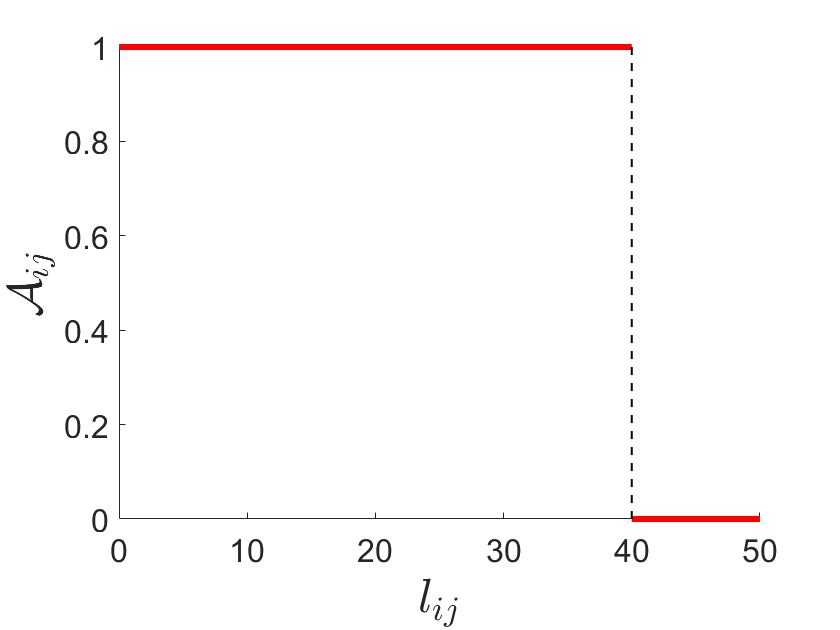}}
        \hfill
  \subfloat[]{%
        \includegraphics[width=0.4\linewidth,trim={0cm 0cm 1cm 1cm},clip]{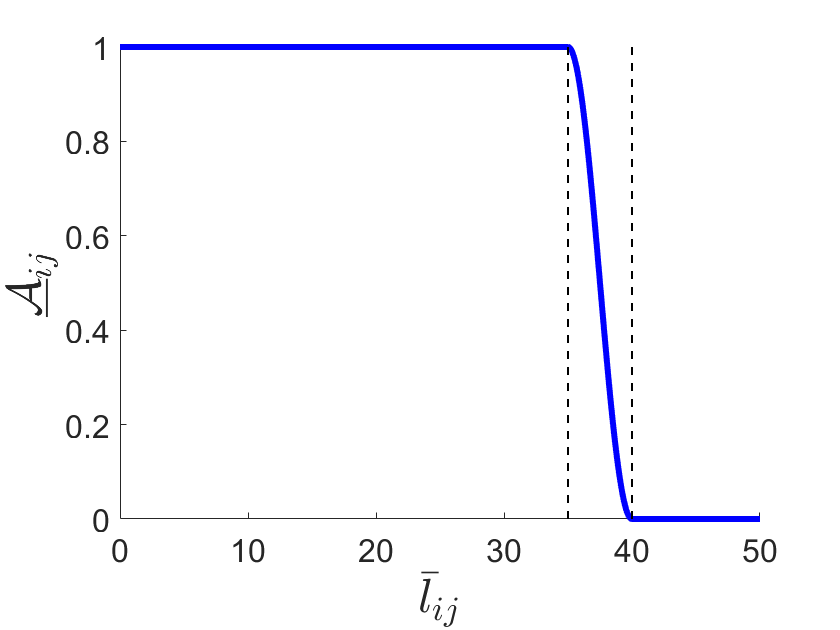}}
        \hfill
  \caption{Edge weights between two robots assuming a communication range of $\Delta = \SI{40}{\meter}$. (a) The binary edge weight $\mathcal{A}_{ij}$ in~(\ref{eqn:binary-edge-weight}) is defined as $\mathcal{A}_{ij} = 1$ if robots $i$ and $j$ are connected, else $\mathcal{A}_{ij} = 0$. (b) For our proposed weighted undirected graph, we define a non-binary edge weight $\underline{\mathcal{A}}_{ij}$ in~(\ref{eqn:conservative-edge-weight}) that gradually goes to $0$ as the distance measure $\bar{l}_{ij}$ goes from $\Delta_0 = \SI{35}{\meter}$ to $\Delta = \SI{40}{\meter}$.}
  \label{fig:edge-weights}
  \vspace{-0.5cm}
\end{figure}

Given the communication range of $\Delta$ between two robots, we introduce a new parameter $\Delta_0$, such that $0~<~\Delta_0~<~\Delta$. Based on the edge weight defined in~\cite{robuffo2013passivity}, we use $\Delta_0$ to define a non-binary edge weight between two robots $i$ and $j$ as:
\vspace{-0.1cm}
\begin{equation}
    \label{eqn:conservative-edge-weight}
    \underline{\mathcal{A}}_{ij} = \begin{cases} \ \ \ \ \ \ \ \ \ \ \ 1 & 0 \leq \bar{l}_{ij} \leq \Delta_0 \\
\frac{1}{2} + \frac{1}{2}\cos{ \left [ \frac{\pi(\bar{l}_{ij} - \Delta_0)}{\Delta - \Delta_0}   \right ] } & \Delta_0 < \bar{l}_{ij} \leq \Delta \\
\ \ \ \ \ \ \ \ \ \ \ 0 & \bar{l}_{ij} > \Delta
\end{cases}.
\end{equation}
Fig.~\ref{fig:edge-weights} compares $\underline{\mathcal{A}}_{ij}$ with $\mathcal{A}_{ij}$ from~(\ref{eqn:binary-edge-weight}). We then proceed to define the corresponding degree matrix $\underline{\mathcal{D}}~=~\texttt{diag}[\underline{d}_i]$ with $\underline{d}_i = \sum^{n}_{j=1} \underline{\mathcal{A}}_{ij}$ and the corresponding Laplacian matrix $\underline{\mathcal{L}} = \underline{\mathcal{D}} - \underline{\mathcal{A}}$. Finally, we use the algebraic connectivity of this weighted undirected graph $\underline{\lambda}^{\mathcal{L}}_2$ as an indicator for the connectivity of the system with uncertain robot positions.

Next, we proceed to derive the value of $\delta_\mathcal{E}$ required in~(\ref{eqn:confidence-ellipse}). We define the following events:
\vspace{-0.1cm}
\begin{align*}
    \mathbb{E}_1:\ \ & \mathbf{p}_i \in \mathcal{S}_i \ \forall \ i \in [1,N], \\
    \mathbb{E}_2:\ \  & \bar{l}_{ij} \geq {l}_{ij} \ \forall \ \left \{ i,j \right \} \in [1,N] \times [1,N] \mid i \neq j, \\ 
    \mathbb{E}_3:\ \  & \underline{\mathcal{A}}_{ij} \leq {\mathcal{A}}_{ij} \ \forall \ \left \{ i,j \right \} \in [1,N] \times [1,N] \mid i \neq j, \\
    \mathbb{E}_4:\ \  & \lambda_2^{\mathcal{L}} \geq \underline{\lambda}_2^{\mathcal{L}}.
\end{align*}
Here $\mathbb{E}_1$ represents the event that all robot positions lie within their corresponding circular regions. We assume that the true positions $\mathbf{p}_{i}$ of the robots in the system are independent of each other. Thus, using~(\ref{eqn:circular-region}) we express the probability of event $\mathbb{E}_1$ as:
\vspace{-0.1cm}
\begin{equation}
    \label{eqn:prob-E1-deltaE}
    \Pr[\mathbb{E}_1] = \prod_{i = 1}^{N}\Pr[\mathbf{p}_i \in \mathcal{S}_i] \geq \prod_{i = 1}^{N} (1-\delta_{\mathcal{E}}) = (1-\delta_{\mathcal{E}})^N,
\end{equation}
where $N$ is the number of robots in the system. 

$\mathbb{E}_2$ represents the event that the distance measures $\bar{l}_{ij}$ between any two robots $i$ and $j$ will always be greater than or equal to the true distance ${l}_{ij}$. We proceed to derive the probability of event $\mathbb{E}_2$ as:
\vspace{-0.1cm}
\begin{align}
    \label{eqn:prob-E2-E1}
    \Pr[\mathbb{E}_2] & = \Pr[\mathbb{E}_2 \mid \mathbb{E}_1]\cdot \Pr[\mathbb{E}_1] + \Pr[\mathbb{E}_2 \mid \mathbb{E}'_1]\cdot \Pr[\mathbb{E}'_1] \nonumber \\
    & \geq \Pr[\mathbb{E}_2 \mid \mathbb{E}_1]\cdot \Pr[\mathbb{E}_1] = \Pr[\mathbb{E}_1],
\end{align}
where $\Pr[\mathbb{E}_2 \mid \mathbb{E}_1] = 1$ since for any two robots $i$ and $j$ if $\mathbf{p}_i \in \mathcal{S}_i$ and $\mathbf{p}_j \in \mathcal{S}_j$, then $\bar{l}_{ij} \geq {l}_{ij}$ as shown in Fig.~\ref{fig:conservative-length}. 

$\mathbb{E}_3$ represents the event that the non-binary edge weight $\underline{\mathcal{A}}_{ij}$ from~(\ref{eqn:conservative-edge-weight}) is less than the edge weight ${\mathcal{A}}_{ij}$ from~(\ref{eqn:binary-edge-weight}) for any two robots $i$ and $j$. Similar to~(\ref{eqn:prob-E2-E1}), we derive the probability of event $\mathbb{E}_3$ as:
\vspace{-0.1cm}
\begin{align}
    \label{eqn:prob-E3-E2}
    \Pr[\mathbb{E}_3] & = \Pr[\mathbb{E}_3 \mid \mathbb{E}_2]\cdot \Pr[\mathbb{E}_2] + \Pr[\mathbb{E}_3 \mid \mathbb{E}'_2]\cdot \Pr[\mathbb{E}'_2] \nonumber \\
    & \geq \Pr[\mathbb{E}_3 \mid \mathbb{E}_2]\cdot \Pr[\mathbb{E}_2] = \Pr[\mathbb{E}_2],
\end{align}
where $\Pr[\mathbb{E}_3 \mid \mathbb{E}_2] = 1$ since for any two robots $i$ and $j$ if $\bar{l}_{ij} \geq {l}_{ij}$, then $\underline{\mathcal{A}}_{ij} \leq {\mathcal{A}}_{ij}$ as shown in Fig.~\ref{fig:edge-weights}. 

Finally, $\mathbb{E}_4$ represents the event that the algebraic connectivity of our weighted undirected graph $\underline{\lambda}_2^{\mathcal{L}}$ is less than or equal to the true algebraic connectivity $\lambda_2^{\mathcal{L}}$ \new{(obtained using the adjacency matrix defined in~(\ref{eqn:binary-edge-weight}))}. The probability of event $\mathbb{E}_4$ is derived as:
\vspace{-0.1cm}
\begin{align}
    \label{eqn:prob-E4-E3}
    \Pr[\mathbb{E}_4] & = \Pr[\mathbb{E}_4 \mid \mathbb{E}_3]\cdot \Pr[\mathbb{E}_3] + \Pr[\mathbb{E}_4 \mid \mathbb{E}'_3]\cdot \Pr[\mathbb{E}'_3] \nonumber \\
    & \geq \Pr[\mathbb{E}_4 \mid \mathbb{E}_3]\cdot \Pr[\mathbb{E}_3] = \Pr[\mathbb{E}_3],
\end{align}
where $\Pr[\mathbb{E}_4 \mid \mathbb{E}_3] = 1$ since by definition the algebraic connectivity monotonically increases as the graph edge weights increase~\cite{yang2010decentralized,robuffo2013passivity}. Thus, from~(\ref{eqn:prob-E1-deltaE})-(\ref{eqn:prob-E4-E3}) we have:
\vspace{-0.1cm}
\begin{equation}
    \label{eqn:prob-E4-deltaE}
    \Pr[ \lambda_2^{\mathcal{L}} \geq \underline{\lambda}_2^{\mathcal{L}} ] \geq (1-\delta_{\mathcal{E}})^N,
\end{equation}
which shows that $\underline{\lambda}_2^{\mathcal{L}}$ lower-bounds $\lambda_2^{\mathcal{L}}$ with a minimum probability value of $(1-\delta_{\mathcal{E}})^N$. If the value of $\underline{\lambda}_2^{\mathbb{L}}$ is maintained above the specified lower limit $\epsilon$ from~(\ref{eqn:objective}), i.e., if $\underline{\lambda}_2^{\mathbb{L}} > \epsilon$, then from~(\ref{eqn:prob-E4-deltaE}) we get:
\vspace{-0.1cm}
\begin{equation}
    \label{eqn:prob-min-ac-deltaE}
    \Pr[ \lambda_2^{\mathcal{L}} > \epsilon ] \geq (1-\delta_{\mathcal{E}})^N.
\end{equation}
In order to satisfy the connectivity maintenance requirement described in~(\ref{eqn:connectivity-maintenance}), we set $(1-\delta_{\mathcal{E}})^N = 1-\delta$, which finally gives us the following value for $\delta_{\mathcal{E}}$:
\vspace{-0.1cm}
\begin{equation}
    \label{eqn:deltaE-value}
    \delta_{\mathcal{E}} = 1-(1-\delta)^{(1/N)}, \vspace{-0.02cm}
\end{equation}
where $\delta$ is the probability value representing a desired confidence level in~(\ref{eqn:connectivity-maintenance}). \new{Thus, setting the value of $\delta_{\mathcal{E}}$ as shown in~(\ref{eqn:deltaE-value}) and ensuring that $\underline{\lambda}_2^{\mathcal{L}}$ is maintained above $\epsilon$ results in satisfying the connectivity maintenance requirement from~(\ref{eqn:connectivity-maintenance}).}

\new{Note that the weighted undirected graph is a conservative representation of the system connectivity since it measures the connectivity based on overbounding circular regions $\mathcal{S}_i$ as opposed to true robot positions $\mathbf{p}_i$. Thus, while maintaining $\underline{\lambda}_2^{\mathcal{L}}$ above $\epsilon$ satisfies the connectivity maintenance requirement, it can result in restricting the mobility of the system. Here the amount of restriction depends on the size of the overbounding circular regions $\mathcal{S}_i$, which depend on the amount of uncertainty in the robot motion and sensing models in~(\ref{eqn:motion-model})-(\ref{eqn:measurement-model}).}

\new{While in this paper we evaluate our algorithm in two-dimensions,} the weighted undirected graph defined in this section can be directly extended to three-dimensions. In the three-dimensional case $\mathcal{E}_i$ in~(\ref{eqn:confidence-ellipse}) would represent a confidence ellipsoid for robot $i$ and $\mathcal{S}_i$ in~(\ref{eqn:circular-region}) would represent the corresponding overbounding spherical region.

%% file: trajectory_planning.tex
\label{sec:trajectory-planning}

In this section, we present the details of our trajectory planning algorithm for solving the problem stated in~(\ref{eqn:objective}). First, we define a connectivity cost function based on the algebraic connectivity $\underline{\lambda}^{\mathcal{L}_t}_2$ of the weighted undirected graph defined in Section~\ref{sec:connectivity-metric}. We incorporate this cost function with the cost in~(\ref{eqn:objective}) to obtain a transformed planning problem. Next, we present a distributed ADMM setup in order to solve the transformed problem and plan nominal trajectories for the multi-robot system. We then describe the method used for performing the optimization step within the ADMM setup and analyze the complexity of its computational bottleneck. Finally, we present an approach to reduce the computational load of this optimization step by deriving an approximation for the required Hessian matrices. \new{Later in Section~\ref{sec:simulations} we demonstrate how the proposed planning algorithm can be utilized for connectivity maintenance under real-time constraints.}
\vspace{-0.35cm}
\subsection{Connectivity cost and transformed planning problem}
\label{subsec:connectivity-cost-function}

As discussed earlier in Section~\ref{sec:connectivity-metric}, maintaining $\underline{\lambda}_2^{\mathcal{L}_t}$ above the specified lower limit $\epsilon$ enables us to satisfy the connectivity maintenance requirement described in~(\ref{eqn:connectivity-maintenance}). Thus, in order to maintain $\underline{\lambda}_2^{\mathcal{L}_t}$ above $\epsilon$, we define a connectivity cost function that grows to infinity as $\underline{\lambda}_2^{\mathcal{L}_t}$ approaches $\epsilon$. Various cost functions with the above property have been proposed in related work~\cite{Sabattini2013} and~\cite{robuffo2013passivity}. For a distributed ADMM setup, it has been shown that the ADMM iteration complexity is inversely proportional to the algebraic connectivity of the system~\cite{makhdoumi2017convergence}. Thus, we choose to define the connectivity cost function for any time instant $t$ as following:
\vspace{-0.1cm}
\begin{equation}
    \label{eqn:connectivity-cost}
    J^c_t = \frac{k_c}{(\underline{\lambda}_2^{\mathcal{L}_t} - \epsilon)} \ \ \forall \ \underline{\lambda}_2^{\mathcal{L}_t} > \epsilon,
\end{equation}
where $k_c$ is a parameter that determines the magnitude of the cost function. In order to incorporate the connectivity cost function with~(\ref{eqn:objective}), we update original planning problem as follows:
\vspace{-0.1cm}
\begin{align}
    \label{eqn:cost-function-connectivity}
    \check{U} & = \argmin_{} \left ( \sum_{t = 0}^T \sum_{i = 1}^{N} J_{i,t}[\check{\mathbf{b}}_{i,t},\check{\mathbf{u}}_{i,t}]  + \sum_{t = 0}^T J^c_t \right ) \\
    & = \argmin_{} \sum_{t = 0}^{T} \sum_{i = 1}^{N} \tilde{J}_{i,t},
\end{align}
\vspace{-0.1cm}
where:
\vspace{-0.1cm}
\begin{equation*}
    \tilde{J}_{i,t} = J_{i,t}[\check{\mathbf{b}}_{i,t},\check{\mathbf{u}}_{i,t}] + \frac{1}{N} J^c_t.
\end{equation*}
Thus, we write the transformed planning problem as:
\vspace{-0.1cm}
\begin{equation}
    \begin{gathered}
    \label{eqn:transformed-opt-problem}
    \vspace{-0.2cm} \check{U} = \argmin_{} \sum_{t = 0}^{T} \sum_{i = 1}^{N} \tilde{J}_{i,t} \\
    \text{subject to:} \hspace{5cm} \\
    \check{\mathbf{b}}_{i,0} = {\mathbf{b}}_{i,\text{init}} \ \forall \ i \in [1,N], \\
    \check{\mathbf{b}}_{i,t+1} = \mathbf{g}_i[\check{\mathbf{b}}_{i,t}, \check{\mathbf{u}}_{i,t}] \ \forall \ i \in [1,N], \forall \ t \in [0,T-1].
    \end{gathered}
\end{equation}

The difference between~(\ref{eqn:objective}) and the transformed planning problem is that the connectivity maintenance constraint has been incorporated in the cost function. The main reason behind transforming the planning problem from~(\ref{eqn:objective}) to~(\ref{eqn:transformed-opt-problem}) is to allow us to use existing optimization tools~\cite{van2012motion} within the ADMM setup, as will be discussed in Section~\ref{subsec:bsilqg}. \new{Note that $J^c_t$ in~(\ref{eqn:connectivity-cost}) and consequently $\tilde{J}_{i,t}$ in~(\ref{eqn:transformed-opt-problem}) are undefined for $\underline{\lambda}_2^{\mathcal{L}_t} \leq \epsilon$. In order to avoid numerical instability, we use a line-search method (discussed later in~(\ref{eqn:admm-line-search})) that ensures $\underline{\lambda}_2^{\mathcal{L}_t} > \epsilon$ for the initial guess of the trajectory optimization. If $\underline{\lambda}_2^{\mathcal{L}_t}$ gets close to $\epsilon$ during the optimization process, the nature of the cost function in~(\ref{eqn:connectivity-cost}) results in gradients that drive $\underline{\lambda}_2^{\mathcal{L}_t}$ away from $\epsilon$.}


\vspace{-0.35cm}
\subsection{Distributed ADMM setup}
\label{subsec:admm-setup}

In order to solve the transformed planning problem from~(\ref{eqn:transformed-opt-problem}), we implement a distributed ADMM setup~\cite{Boyd2010} that iteratively plans nominal trajectories for the multi-robot system. In each ADMM iteration, each robot optimizes only a subset of the robot trajectories in order to reduce the computational load of the optimization step. The optimized trajectories are then communicated with the rest of the system. After the communication step, each robot updates its local ADMM consensus and dual variables before moving on to the next ADMM iteration. When the stopping criteria is satisfied, the last updated local ADMM consensus variable is used as the planned nominal trajectories for the multi-robot system.

Each robot $i$ begins by generating an initial guess for the nominal trajectories of the multi-robot system $\check{U}^{(i,1)}$, where the superscript denotes that the variable is stored locally on robot $i$ and is for the first ADMM iteration. The initial guess is typically generated based on the local tasks for each robot. We assume that the process used to generate the initial guess maintains $\underline{\lambda}_2^{\mathcal{L}_t}$ above $\epsilon \ \forall \ t \in [0,T]$. Later in Section~\ref{subsec:sim-setup}, we describe our method for obtaining the initial guess when the local task for each robot involves reaching a desired position. Once the initial guess has been generated, the robot proceeds to initialize its local copy of the consensus variable as $\bar{U}^{(i,1)} = \check{U}^{(i,1)}$. The ADMM dual variable $Y^{(i,1)}$ is initialized as a zero matrix.

Next, the robot begins the ADMM iterations. In each ADMM iteration $k$, the robot first obtains a subset $\mathcal{V}^{(i,k)}$ containing indices of the robot trajectories to optimize. Different strategies can be deployed for obtaining $\mathcal{V}^{(i,k)}$. For example, setting $\mathcal{V}^{(i,k)} = \{i\}$ results in a greedy optimization where the robot optimizes its own trajectory; setting $\mathcal{V}^{(i,k)}$ to contain neighboring robot indices focuses more on the local connectivity rather than the global connectivity of the system. In our algorithm, we obtain $\mathcal{V}^{(i,k)}$ such that it contains $i$ and cycles through the indices of \new{the other $(N-1)$ robots. As mentioned in Section~\ref{subsec:trajecory-planning}, we assume that each robot knows there are $N$ number of robots in the system}. Let $\eta$ represent the number of elements in $\mathcal{V}^{(i,k)}$. Table~\ref{tab:optimization-subset} shows an example of the subsets $\mathcal{V}^{(i,k)}$ for four ADMM iterations in a system with four robots and with $\eta = 3$. We observe that this strategy for obtaining $\mathcal{V}^{(i,k)}$ avoids the problem of greedy optimizations and eventually results in nominal trajectories for the system with lower overall costs as shown in Section~\ref{sec:simulations}. \new{The value of $\eta$ can be chosen based on the computation power of each robot. While choosing a larger $\eta$ would result in lower overall costs, the computational load would be higher.}
\vspace{-0.15cm}

\begin{table}[h]
    \centering
\begin{tabular}{ |c|c|c|c|c| } 
 \hline
  & $\mathcal{V}^{(1,k)}$ & $\mathcal{V}^{(2,k)}$ & $\mathcal{V}^{(3,k)}$ & $\mathcal{V}^{(4,k)}$ \\
 \hline
 $k=1$ & $\{1,2,3\}$ & $\{2,3,4\}$ & $\{3,4,1\}$ & $\{4,1,2\}$ \\
 \hline
 $k=2$ & $\{1,3,4\}$ & $\{2,4,1\}$ & $\{3,1,2\}$ & $\{4,2,3\}$ \\
 \hline
 $k=3$ & $\{1,4,2\}$ & $\{2,1,3\}$ & $\{3,2,4\}$ & $\{4,3,1\}$ \\
 \hline
 $k=4$ & $\{1,2,3\}$ & $\{2,3,4\}$ & $\{3,4,1\}$ & $\{4,1,2\}$ \\
 \hline
\end{tabular}
    \caption{Example of subsets $\mathcal{V}$ for a system with four robots, where $\eta = 3$ and up to $k=4$ ADMM iterations are considered.}
    \label{tab:optimization-subset}
    \vspace{-0.25cm}
\end{table}

Once the subset $\mathcal{V}^{(i,k)}$ has been obtained, the robot performs the optimization step. In this step, we first initialize $\check{U}^{(i,k+1)}~=~\bar{U}^{(i,k)}$ and then only update the trajectories for robots in subset $\mathcal{V}^{(i,k)}$. Based on the cost function in~(\ref{eqn:transformed-opt-problem}), the robot optimizes the following augmented cost~\cite{Boyd2010} to obtain optimized trajectories for robots in $\mathcal{V}^{(i,k)}$:
\begin{equation}
\begin{split}
    \label{eqn:admm-opt}
    \check{U}^{(i,k+1)}_{\mathcal{V}^{(i,k)}} & =  \argmin_{\check{U}_{\mathcal{V}^{(i,k)}}} \sum_{t=0}^T \Bigg\{ \sum_{j \in \mathcal{V}^{(i,k)}} \tilde{J}_{j,t} \\
    & + \mathbf{y}^{(i,k)\top}_{\mathcal{V}^{(i,k)},t} \left ( \check{\mathbf{u}}_{\mathcal{V}^{(i,k)},t} - \bar{\mathbf{u}}^{(i,k)}_{\mathcal{V}^{(i,k)},t} \right ) \\ 
    & + (\rho/ 2) \left \| \check{\mathbf{u}}_{\mathcal{V}^{(i,k)},t} - \bar{\mathbf{u}}^{(i,k)}_{\mathcal{V}^{(i,k)},t} \right \|_2^2 \Bigg\},
\end{split}
\end{equation}
\vspace{-0.8cm}
\begin{align}
    \text{subject to:} \hspace{1cm} \nonumber \\
    \check{\mathbf{b}}_{i,0} = & \ \mathbf{b}_{i,\text{init}} \ \forall \ i \in [1,N], \nonumber \\
    \check{\mathbf{b}}_{i,t+1} = \mathbf{g}_i[\check{\mathbf{b}}_{i,t}, \check{\mathbf{u}}_{i,t} &] \ \forall \ i \in [1,N], \forall \ t \in [0,T-1], \nonumber
\end{align}
where $\rho > 0$ is the ADMM penalty weight, $\check{\mathbf{u}}$ and $\bar{\mathbf{u}}^{(i,k)}$ represent the corresponding vectors from matrices $\check{U}$ and $\bar{U}^{(i,k)}$ respectively, and $\mathbf{y}^{(i,k)}$ represents the corresponding vector from dual variable matrix $Y^{(i,k)}$. We discuss the method used to solve this optimization step later in Section~\ref{subsec:bsilqg}. \new{Here the cost accompanied by $\rho$ is commonly referred to as the consensus constraint~\cite{Boyd2010}. Since each robot optimizes trajectories for a subset of robots, each robot might have a different idea of the planned nominal trajectories for the complete system. Thus, enforcing this constraint allows the robots reach a consensus on the planned nominal trajectories.}

After the optimization step, the robot proceeds to the communication step where each robot $i$ shares the optimized trajectories $\check{U}^{(i,k+1)}_{\mathcal{V}^{(i,k)}}$ obtained using~(\ref{eqn:admm-opt}). In this step, each robot receives the optimized trajectories from all other robots in the system, potentially via  multi-hop communication. Note that our planner is distributed since each robot optimizes with respect to a subset of trajectories, \new{as opposed to trajectories from all robots in the system. However, the communication architecture is centralized since each robot shares information with all other robots in the system. Thus, the communication load does not scale well with the number of robots and could lead to delays during execution. Later in Section~\ref{subsec:online-planning} we account for a communication delay while evaluating our planner under real-time constraints.}


Once the communication step is complete, each robot $i$ receives the optimized trajectories from all other robots. Note that the trajectory for each robot has been optimized $\eta$ times across the system. For example, in Table~\ref{tab:optimization-subset} at ADMM iteration $k=3$, the trajectory for robot $4$ was optimized by robots $1, 3$ and $4$, i.e., $\eta = 3$ times. Thus, based on the consensus update step in~\cite{Boyd2010}, the robot calculates an average optimized trajectory for each robot $j$ as follows:
\vspace{-0.1cm}
\begin{equation}
    \label{eqn:admm-avg}
    \tilde{U}^{(i,k+1)}_j = \frac{1}{\eta}\sum_{l=1}^{N} \check{U}^{(l,k+1)}_j \cdot \mathds{1}_{\mathcal{V}^{(l,k)}}[j],
\end{equation}
where $\mathds{1}_{\mathcal{V}^{(l,k)}}[j]$ is an indicator function equal to $1$ if $j \in \mathcal{V}^{(l,k)}$ and equal to $0$ otherwise. \new{Note that in~(\ref{eqn:admm-avg}) the robot does not need to keep track of which robot it received the optimized trajectory $\check{U}^{(l,k+1)}_j$ from during the communication step.}


After the averaging step, it is possible that $\tilde{U}^{(i,k+1)}$ might result in a trajectory for the multi-robot system that does not maintain $\underline{\lambda}_2^{\mathcal{L}_t}$ above the specified lower limit of $\epsilon$. Thus, in order to ensure that the consensus variable $\bar{U}$ always results in trajectories that maintain $\underline{\lambda}_2^{\mathcal{L}_t} > \epsilon$, we use a line search algorithm~\cite{more1994line} to update $\bar{U}$. We limit the change to $\bar{U}$ between iterations as follows:
\vspace{-0.1cm}
\begin{equation}
    \label{eqn:admm-line-search}
    \bar{U}^{(i,k+1)} = \bar{U}^{(i,k)} + \beta \cdot ( \tilde{U}^{(i,k+1)} - \bar{U}^{(i,k)}),
\end{equation}
where $\beta$ is a parameter that determines the amount of change in $\bar{U}$. We begin with $\beta = 1$ and check if the corresponding $\bar{U}^{(i,k+1)}$ results in trajectories that maintain $\underline{\lambda}_2^{\mathcal{L}_t} > \epsilon$. If $\underline{\lambda}_2^{\mathcal{L}_t}$ is not maintained above $\epsilon$, we reduce $\beta$ by a factor $\gamma$ as: $\beta = \gamma \cdot \beta$, where $0 < \gamma < 1$. We then calculate the new $\bar{U}^{(i,k+1)}$ using~(\ref{eqn:admm-line-search}) and repeat the process until $\bar{U}^{(i,k+1)}$ results in trajectories that maintain $\underline{\lambda}_2^{\mathcal{L}_t} > \epsilon$. \new{Thus, if $\tilde{U}^{(i,k+1)}$ obtained from the averaging step does not maintain $\underline{\lambda}_2^{\mathcal{L}_t} > \epsilon$, the line search algorithm gradually scales the trajectory to $\bar{U}^{(i,k)}$. Since we assume that the initial nominal trajectory guess $\bar{U}^{(i,0)}$} maintains $\underline{\lambda}_2^{\mathcal{L}_t} > \epsilon$, the line search in~(\ref{eqn:admm-line-search}) ensures that $\bar{U}$ always results in trajectories that maintain $\underline{\lambda}_2^{\mathcal{L}_t} > \epsilon$. \new{Note that each robot performs the line search algorithm with the same value of $\gamma$ (specified later in Section~\ref{sec:simulations}). Thus, the updated consensus variable $\bar{U}^{(i,k+1)}$ is the same on all robots.}


Finally, each robot $i$ updates its ADMM dual variable as follows:
\vspace{-0.1cm}
\begin{equation}
    \label{eqn:admm-dual-update}
    Y^{(i,k+1)} = Y^{(i,k)} + \rho \cdot ( \check{U}^{(i,k+1)} - \bar{U}^{(i,k+1)} ),
\end{equation}
where $\rho$ is the ADMM penalty weight defined in~(\ref{eqn:admm-opt}).

Before beginning the next ADMM iteration, the robot checks if the stopping criterion has been satisfied. \new{The stopping criteria can be either convergence-based or time-based. Later in Section~\ref{sec:simulations} we evaluate our planner under a time-based stopping criterion.} If the stopping criteria is satisfied, the robots set the last updated value of $\bar{U}$ as the planned nominal trajectories for the multi-robot system. Since $\bar{U}$ always results in trajectories that maintain $\underline{\lambda}_2^{\mathcal{L}_t} > \epsilon$, the output from our trajectory planning algorithm always results in trajectories that maintain $\underline{\lambda}_2^{\mathcal{L}_t} > \epsilon$. Thus, our algorithm satisfies the connectivity maintenance requirement from~(\ref{eqn:connectivity-maintenance}). Algorithm~\ref{alg:admm} summarizes our trajectory planning algorithm. 

\begin{algorithm}[t]
\caption{Trajectory planner}\label{alg:admm}
\begin{algorithmic}[1]
\For{$i$ = $1,\dots,N$} in parallel
    \State{Generate initial nominal trajectory guess $\check{U}^{(i,1)}$}
    \State{Initialize consensus variable $\bar{U}^{(i,1)} = \check{U}^{(i,1)}$, dual}
    \Statex variable $Y^{(i,1)}$ as zero matrix, and ADMM iteration $k=1$
    \While{stopping criterion is not satisfied}
    \State{Obtain subset $\mathcal{V}^{(i,k)}$ of trajectories to optimize}
    \State{Perform the optimization step (Equation (\ref{eqn:admm-opt})) to}
    \Statex \hspace{0cm}obtain $\check{U}^{(i,k+1)}_{\mathcal{V}^{(i,k)}}$
    \State{Communicate $\check{U}^{(i,k+1)}_{\mathcal{V}^{(i,k)}}$ to (and from) other robots}
    \State{Calculate average optimized trajectories (Equation}
    \Statex \hspace{0cm}(\ref{eqn:admm-avg})) to obtain $\tilde{U}^{(i,k+1)}$
    \State{Update consensus variable $\bar{U}^{(i,k+1)}$ using line}
    \Statex \hspace{0cm}search algorithm (Equation (\ref{eqn:admm-line-search}))
    \State{Update dual variable $Y^{(i,k+1)}$ (Equation (\ref{eqn:admm-dual-update}))}
    \State{Update ADMM iteration $k=k+1$}
    \EndWhile
    \State{Set planned nominal trajectories as $\check{U} = \bar{U}^{(i,k)}$}
\EndFor
\end{algorithmic}
\end{algorithm}

\vspace{-0.35cm}
\subsection{ADMM trajectory optimization and complexity analysis}
\label{subsec:bsilqg}
In order to obtain the optimized nominal trajectories $\check{U}^{(i,k+1)}_{\mathcal{V}^{(i,k)}}$ in~(\ref{eqn:admm-opt}), we use the belief-space iterative Linear Quadratic Gaussian (belief-space iLQG) method~\cite{van2012motion}. Since the analysis in the remainder of this section is applicable for a general subset of robot trajectories, we simply represent $\mathcal{V}^{(i,k)}$ as $\mathcal{V}$. We first extend~(\ref{eqn:belief-dynamics}) to define concatenated belief dynamics for the subset $\mathcal{V}$ as follows:
\vspace{-0.1cm}
\begin{equation}
    \label{eqn:concatenated-belief-dynamics}
    \mathbf{b}_{\mathcal{V},t+1} = \mathbf{g}_{\mathcal{V}}[\mathbf{b}_{\mathcal{V},t}, \mathbf{u}_{\mathcal{V},t}] + M_{\mathcal{V}}[\mathbf{b}_{\mathcal{V},t}, \mathbf{u}_{\mathcal{V},t}]\mathbf{m}_{\mathcal{V},t},
\end{equation}
where $\mathbf{b}_{\mathcal{V}}$ is the concatenated belief vector of robots in the subset $\mathcal{V}$. Next, similar to~(\ref{eqn:nominal-dynamics}), the concatenated nominal trajectory for the subset $\mathcal{V}$ is represented as $( \check{\mathbf{b}}_{\mathcal{V},0}, \check{\mathbf{u}}_{\mathcal{V},0}, \dots , \check{\mathbf{b}}_{\mathcal{V},T-1}, \check{\mathbf{u}}_{\mathcal{V},T-1}, \check{\mathbf{b}}_{\mathcal{V},T} )$, such that:
\vspace{-0.1cm}
\begin{equation}
    \label{eqn:concatenated-nominal-dynamics}
    \check{\mathbf{b}}_{\mathcal{V},t+1} = \mathbf{g}_{\mathcal{V}}[\check{\mathbf{b}}_{\mathcal{V},t}, \check{\mathbf{u}}_{\mathcal{V},t}]\ \forall \ t \in [0,T-1].
\end{equation}
Additionally, we rewrite the ADMM optimization step in~(\ref{eqn:admm-opt}) as:
\vspace{-0.1cm}
\begin{equation}
    \begin{gathered}
    \label{eqn:admm-cost-c}
    \check{U}^{(i,k+1)}_{\mathcal{V}} = \argmin_{\check{U}_{\mathcal{V}}} \sum_{t=0}^T c_t[\check{\mathbf{b}}_{\mathcal{V},t}, \check{\mathbf{u}}_{\mathcal{V},t}], \\
    \text{subject to:} \hspace{5cm} \\
    \check{\mathbf{b}}_{\mathcal{V},0} = {\mathbf{b}}_{\mathcal{V},\text{init}}, \\
    \check{\mathbf{b}}_{\mathcal{V},t+1} = \mathbf{g}_\mathcal{V}[\check{\mathbf{b}}_{\mathcal{V},t}, \check{\mathbf{u}}_{\mathcal{V},t}] \ \forall \ t \in [0,T-1],
    \end{gathered}
\end{equation}
where $c_t$ is the cost at time instant $t$. While $c_t$ depends on the belief and input vectors of the entire multi-robot system, for simplicity we write $c_t$ to be a function of only $\check{\mathbf{b}}_{\mathcal{V}}$ and $\check{\mathbf{u}}_{\mathcal{V}}$ since only the trajectories for subset $\mathcal{V}$ are being optimized. The belief-space iLQG method~\cite{van2012motion} begins with an initial guess for the nominal trajectory and computes a locally optimal solution for~(\ref{eqn:admm-cost-c}) by performing backward value iteration. The value iteration process involves quadratizing the cost function $c_t$ along the nominal trajectory as follows:
\begin{equation}
    \label{eqn:bsilqg-cost-approx}
    c_t \approx \frac{1}{2} \begin{bmatrix} \delta \mathbf{b}\\\delta \mathbf{u} \end{bmatrix}^\top \begin{bmatrix} \check{c}_{\mathbf{bb},t} & \check{c}_{\mathbf{bu},t}^\top \\ \check{c}_{\mathbf{bu},t} & \check{c}_{\mathbf{uu},t} \end{bmatrix} \begin{bmatrix} \delta \mathbf{b}\\\delta \mathbf{u} \end{bmatrix} + \begin{bmatrix} \delta \mathbf{b}\\\delta \mathbf{u} \end{bmatrix}^\top \begin{bmatrix} \check{\mathbf{c}}_{\mathbf{b},t} \\ \check{\mathbf{c}}_{\mathbf{u},t} \end{bmatrix} + \check{c}_t,
\end{equation}
where:
\begin{alignat*}{4}
  &\delta \mathbf{b} &&= \mathbf{b}_{\mathcal{V},t} - \check{\mathbf{b}}_{\mathcal{V},t},\ &&\delta \mathbf{u} &&= \mathbf{u}_{\mathcal{V},t} - \check{\mathbf{u}}_{\mathcal{V},t},\\
  &\check{c}_{\mathbf{bb},t}\hspace{-0.05cm}&&=\hspace{-0.05cm}\frac{\partial^2 c_t }{ \partial\mathbf{b}_\mathcal{V} \partial\mathbf{b}_\mathcal{V} } [\check{\mathbf{b}}_{\mathcal{V},t}, \check{\mathbf{u}}_{\mathcal{V},t}], \  &&\check{c}_{\mathbf{bu},t}\hspace{-0.05cm}&&=\hspace{-0.05cm}\frac{\partial^2 c_t }{ \partial\mathbf{b}_\mathcal{V} \partial\mathbf{u}_\mathcal{V} } [\check{\mathbf{b}}_{\mathcal{V},t}, \check{\mathbf{u}}_{\mathcal{V},t}], \\
  &\check{c}_{\mathbf{uu},t}\hspace{-0.05cm}&&=\hspace{-0.05cm}\frac{\partial^2 c_t }{ \partial\mathbf{u}_\mathcal{V} \partial\mathbf{u}_\mathcal{V} } [\check{\mathbf{b}}_{\mathcal{V},t}, \check{\mathbf{u}}_{\mathcal{V},t}], \ &&\check{\mathbf{c}}_{\mathbf{b},t} &&= \frac{\partial c_t }{ \partial\mathbf{b}_\mathcal{V} } [\check{\mathbf{b}}_{\mathcal{V},t}, \check{\mathbf{u}}_{\mathcal{V},t}], \\
  &\check{\mathbf{c}}_{\mathbf{u},t} &&= \frac{\partial c_t }{ \partial\mathbf{u}_\mathcal{V} } [\check{\mathbf{b}}_{\mathcal{V},t}, \check{\mathbf{u}}_{\mathcal{V},t}],\ &&\check{c}_{t} &&= c_t[\check{\mathbf{b}}_{\mathcal{V},t}, \check{\mathbf{u}}_{\mathcal{V},t}].
\end{alignat*}
As discussed in previous works related to belief-space iLQG~\cite{van2012motion,park2019distributed}, one of the primary sources of a computational bottleneck lies in the computation of the Hessian $\check{c}_{\mathbf{bb},t}$. In our trajectory planner, from~(\ref{eqn:cost-function-connectivity})-(\ref{eqn:admm-opt}) and~(\ref{eqn:admm-cost-c}), note that the cost function $c_t$ consists of the connectivity cost function $J^c_t$. Thus, computation of $\check{c}_{\mathbf{bb},t}$ involves computing the Hessian of the connectivity cost function $\check{J}^c_{\mathbf{bb},t} = \frac{\partial^2 J^c_t} {\partial\mathbf{b}_{\mathcal{V}}\partial \mathbf{b}_{\mathcal{V}}}[ \underline{\lambda}_2^{\mathcal{L}_t} [ \check{\mathbf{b}}_{\mathcal{V},t}]]$. For our belief-space iLQG implementation, we observe that computing $\check{J}^c_{\mathbf{bb},t}$ is the primary computational bottleneck. Thus, we analyze its complexity below. 

For simplicity, we assume that the state vector $\mathbf{x}_{i,t}$ has the same dimension $n$ for all robots in the system throughout the planning horizon. In this case, the dimension of the belief vector for each robot, as defined in~(\ref{eqn:belief-vec}), is $O[n^2]$ since it contains elements from the state estimation covariance matrix. Thus, the dimension of the concatenated belief vector $\mathbf{b}_\mathcal{V}$ is $O[\eta n^2]$, since $\mathcal{V}$ contains $\eta$ elements. Given the dimension of $\mathbf{b}_\mathcal{V}$, the Hessian $\check{J}^c_{\mathbf{bb},t}$ contains $O[\eta^2 n^4]$ entries. The typical approach to compute the required Hessian in previous belief-space iLQG implementations is to use numerical differentiation (central differences)~\cite{van2012motion,park2019distributed}. Using numerical differentiation would require $O[\eta^2 n^4]$ evaluations of $J^c_t$, and consequently $O[\eta^2 n^4]$ evaluations of $\underline{\lambda}_2^{\mathcal{L}_t}$. Considering the entire planning horizon, this results in $O[\eta^2 n^4 T]$ evaluations of $\underline{\lambda}_2^{\mathcal{L}_t}$ per iteration of belief-space iLQG. 

Evaluating $\underline{\lambda}^{\mathcal{L}_t}_2$ requires obtaining the Laplacian matrix $\underline{\mathcal{L}}_t$ whose elements depend on the distance measure as shown in~(\ref{eqn:conservative-edge-weight}). For obtaining the distance measures between all robots in the system, we need to perform eigendecompositions of the covariance matrices $\Sigma_{i,t} \ \forall \ i \in [1,N]$ as shown in~(\ref{eqn:conservative-dist}). Assuming the dimension of the position vector $\mathbf{p}_{i,t}$ to be $\varrho$, each eigendecomposition can be evaluated in $O[\varrho^3]$ time~\cite{pan1999complexity}. Thus, the complexity to obtain $\underline{\mathcal{L}}_t$ is of $O[\varrho^3 N]$. Once we obtain $\underline{\mathcal{L}}_t$, we need to perform another eigendecomposition with complexity of $O[N^3]$ to obtain $\underline{\lambda}^{\mathcal{L}_t}_2$. Thus, the complexity of a single evaluation of $\underline{\lambda}^{\mathcal{L}_t}_2$ is of $O[\max[ \varrho^3 N, N^3 ]]$. Given that we need $O[\eta^2 n^4 T]$ evaluations of $\underline{\lambda}_2^{\mathcal{L}_t}$, we finally have a complexity of $O[\eta^2 n^4 T \cdot \max[ \varrho^3 N, N^3 ] ]$ per iteration of belief-space iLQG. 

Since the belief-space iLQG method is used for the optimization step within each ADMM iteration, using numerical differentiation to compute $\check{J}^c_{\mathbf{bb},t}$ results in a prohibitively large computational load. Thus, in the next subsection we present an approach to approximate $\check{J}^c_{\mathbf{bb},t}$ and consequently reduce the required computational load for the belief-space iLQG method.


\vspace{-0.35cm}
\subsection{Hessian approximation for complexity reduction}
\label{subsec:complexity-reduction}

In this subsection, we drop the time notation for simplicity since the presented approximation is applicable $\forall \ t \in [0,T]$. As discussed in Section~\ref{subsec:bsilqg}, the primary computational bottleneck in our implementation of belief-space iLQG arises in computing $\check{J}^c_{\mathbf{bb},t}$. Thus, in this subsection we derive an analytical expression to approximate $\check{J}^c_{\mathbf{bb},t}$ and show that it significantly reduces the required computational load.

We begin by obtaining the gradient of our metric $\underline{\lambda}_2^{\mathcal{L}}$ with respect to the belief vector $\mathbf{b}_i$ as follows~\cite{yang2010decentralized}:
\vspace{-0.1cm}
\begin{equation}
    \label{eqn:cost-gradient-2}
    \frac{\partial \underline{\lambda}_2^{\mathcal{L}}}{\partial \mathbf{b}_i} = (\underline{\mathbf{e}}_2^{\mathcal{L}})^\top \frac{\partial {\underline{\mathcal{L}}}}{\partial \mathbf{b}_i} (\underline{\mathbf{e}}_2^{\mathcal{L}}) = \sum_{j = 1}^N \frac{\partial \underline{\mathcal{A}}_{ij}}{\partial \mathbf{b}_i} \left (\underline{e}_2^{\mathcal{L},(i)} - \underline{e}_2^{\mathcal{L},(j)}  \right )  ^2,
\end{equation}
where $\underline{\mathbf{e}}_2^{\mathcal{L}}$ is the eigenvector of $\underline{\mathcal{L}}$ corresponding to the eigenvalue $\underline{\lambda}_2^{\mathcal{L}}$, and $\underline{e}_2^{\mathcal{L},(i)}$ is the $i^{th}$ element of $\underline{\mathbf{e}}_2^{\mathcal{L}}$. From~(\ref{eqn:conservative-edge-weight}), we obtain the gradient of $\underline{\mathcal{A}}_{ij}$ with respect to the belief vector $\mathbf{b}_i$ as:
\begin{equation}
    \label{eqn:cost-gradient-3}
    \frac{\partial \underline{\mathcal{A}}_{ij}}{\partial \mathbf{b}_i} = -\frac{\pi}{2(\Delta - \Delta_0)} \sin{ \left [ \frac{\pi(\bar{l}_{ij} - \Delta_0)}{\Delta-\Delta_0} \right ] } \frac{\partial \bar{l}_{ij}}{\partial \mathbf{b}_i}.
\end{equation}
Note that while the belief vector $\mathbf{b}_i$ in~(\ref{eqn:belief-vec}) contains the state estimate and the estimation covariance, the distance measure $\bar{l}_{ij}$ in~(\ref{eqn:conservative-dist}) depends only on the position estimate $\hat{\mathbf{p}}_i$ and the position estimation covariance $\Sigma_i$. Thus, in order to obtain $\frac{\partial \bar{l}_{ij}}{\partial \mathbf{b}_i}$ in~(\ref{eqn:cost-gradient-3}), we only need the gradient of $\bar{l}_{ij}$ with respect to each element of $\hat{\mathbf{p}}_i$ and with respect to each element of $\Sigma_i$. From~(\ref{eqn:conservative-dist}), the gradient of $\bar{l}_{ij}$ with respect to $\hat{p}^{(m)}_i$, i.e., the $m^{th}$ element of $\hat{\mathbf{p}}_i$, is computed as:
\begin{equation}
    \label{eqn:cost-gradient-4}
    \frac{\partial \bar{l}_{ij}}{\partial \hat{p}^{(m)}_i} = \frac{ \left (\hat{p}^{(m)}_i - \hat{p}^{(m)}_j   \right )}{ \left \| \hat{\mathbf{p}}_i - \hat{\mathbf{p}}_j \right \|^2_2 },
\end{equation}
and the gradient of $\bar{l}_{ij}$ with respect to element $(m,b)$ of $\Sigma_i$ is computed as~\cite{stump2008connectivity}:
\begin{equation}
    \label{eqn:cost-gradient-5}
    \frac{\partial \bar{l}_{ij}}{\partial \Sigma^{(m,b)}_i} = \left( \frac{s}{2\sqrt{\bar{\lambda}^{\Sigma_i}}} \right) \texttt{tr} \left[ \left( \frac{\partial \bar{\lambda}^{\Sigma_i}}{\partial \Sigma_i} \right)^\top \left( \frac{\partial \Sigma_i}{\partial \Sigma_i^{(m,b)} } \right) \right],
\end{equation}
where $\texttt{tr}[\cdot]$ represents the trace of a matrix. Furthermore, from~\cite{stump2008connectivity}, we get:
\begin{equation}
    \label{eqn:cost-gradient-6}
    \frac{\partial \bar{\lambda}^{\Sigma_i}}{\partial \Sigma_i} = \frac{(\bar{\mathbf{e}}^{\Sigma_i})(\bar{\mathbf{e}}^{\Sigma_i})^\top}{(\bar{\mathbf{e}}^{\Sigma_i})^\top(\bar{\mathbf{e}}^{\Sigma_i}) },
\end{equation}
where $\bar{\mathbf{e}}^{\Sigma_i}$ is the eigenvector of $\Sigma_i$ corresponding to the largest eigenvalue $\bar{\lambda}^{\Sigma_i}$. Thus,~(\ref{eqn:cost-gradient-5}) simplifies to:
\begin{equation}
    \label{eqn:cost-gradient-7}
    \frac{\partial \bar{l}_{ij}}{\partial \Sigma^{(m,b)}_i} = \left( \frac{s}{2\sqrt{\bar{\lambda}^{\Sigma_i}}} \right) \bar{e}^{\Sigma_i,(m)} \bar{e}^{\Sigma_i,(b)}.
\end{equation}
Equations~(\ref{eqn:cost-gradient-4}) and~(\ref{eqn:cost-gradient-7}) allow us to construct the gradient $\frac{\partial \bar{l}_{ij}}{\partial \mathbf{b}_i}$, which is required to obtain $\frac{\partial \underline{\lambda}_2^{\mathcal{L}}}{\partial\mathbf{b}_i}$ using~(\ref{eqn:cost-gradient-2}) and~(\ref{eqn:cost-gradient-3}). By concatenating the gradients $\frac{\partial \underline{\lambda}_2^{\mathcal{L}}}{\partial\mathbf{b}_i} \ \forall \ i \in \mathcal{V}$, we obtain the gradient $\frac{\partial \underline{\lambda}_2^{\mathcal{L}}}{\partial\mathbf{b}_\mathcal{V}}$.

Next, in order to approximate $\check{J}^c_{\mathbf{bb}}$, we begin by writing the second-order Taylor expansion of $J^c$ about $\check{\mathbf{b}}_{\mathcal{V}}$:
\begin{multline}
    \label{eqn:hessian-approx-1}
    J^c[ \underline{\lambda}_2^{\mathcal{L}} [ \mathbf{b}_\mathcal{V} ] ] \approx \frac{1}{2} \check{J}^c_{\lambda \lambda} ( \underline{\lambda}_2^{\mathcal{L}} [ \mathbf{b}_\mathcal{V} ] - \underline{\lambda}_2^{\mathcal{L}} [ \check{\mathbf{b}}_\mathcal{V} ] )^2 \\ + \check{J}^c_{\lambda}( \underline{\lambda}_2^{\mathcal{L}} [ \mathbf{b}_\mathcal{V} ] - \underline{\lambda}_2^{\mathcal{L}} [ \check{\mathbf{b}}_\mathcal{V} ] ) + \check{J}^c,
\end{multline}
where:
\begin{gather*}
    \check{J}^c_{\lambda \lambda} = \frac{\partial^2 J^c}{\partial \underline{\lambda}_2^{\mathcal{L}} \partial \underline{\lambda}_2^{\mathcal{L}} }[ \underline{\lambda}_2^{\mathcal{L}} [\check{\mathbf{b}}_\mathcal{V} ] ], \ \ \check{J}^c_{\lambda} = \frac{\partial J^c}{\partial \underline{\lambda}_2^{\mathcal{L}}}[ \underline{\lambda}_2^{\mathcal{L}} [\check{\mathbf{b}}_\mathcal{V} ] ], \\ \check{J}^c = J^c[ \underline{\lambda}_2^{\mathcal{L}} [\check{\mathbf{b}}_\mathcal{V} ] ].
\end{gather*}
Here $\check{J}^c_{\lambda \lambda}$ is obtained from~(\ref{eqn:connectivity-cost}) as:
\begin{equation}
    \label{eqn:hessian-lambda}
    \check{J}^c_{\lambda \lambda} = \frac{2 k_c}{ ( \underline{\lambda}_2^{\mathcal{L}}[\check{\mathbf{b}}_\mathcal{V}] - \epsilon)^3 }.
\end{equation}
We then approximate the term $( \underline{\lambda}_2^{\mathcal{L}} [ \mathbf{b}_\mathcal{V} ] - \underline{\lambda}_2^{\mathcal{L}} [ \check{\mathbf{b}}_\mathcal{V} ] )$ in~(\ref{eqn:hessian-approx-1}) using a first-order Taylor expansion about $\check{\mathbf{b}}_\mathcal{V}$ as follows:
\begin{equation}
    \label{eqn:hessian-approx-2}
    \underline{\lambda}_2^{\mathcal{L}} [ \mathbf{b}_\mathcal{V} ] - \underline{\lambda}_2^{\mathcal{L}} [ \check{\mathbf{b}}_\mathcal{V} ] \approx (\mathbf{b}_\mathcal{V} - \check{\mathbf{b}}_\mathcal{V})^\top \mathbf{a},
\end{equation}
where $\mathbf{a} = \left( \frac{ \partial \underline{\lambda}_2^{\mathcal{L}}}{ \partial \mathbf{b}_\mathcal{V} }[\check{\mathbf{b}}_\mathcal{V}] \right)^\top$. By substituting~(\ref{eqn:hessian-approx-2}) in~(\ref{eqn:hessian-approx-1}), we get:
\begin{multline}
    \label{eqn:hessian-approx-3}
    J^c[ \underline{\lambda}_2^{\mathcal{L}} [ \mathbf{b}_\mathcal{V} ] ] \approx \frac{1}{2} (\mathbf{b}_\mathcal{V} - \check{\mathbf{b}}_\mathcal{V})^\top ( \check{J}^c_{\lambda \lambda} \mathbf{a} \mathbf{a}^\top ) (\mathbf{b}_\mathcal{V} - \check{\mathbf{b}}_\mathcal{V}) \\ + (\mathbf{b}_\mathcal{V} - \check{\mathbf{b}}_\mathcal{V})^\top (\check{J}^c_{\lambda} \mathbf{a}) + \check{J}^c,
\end{multline}
where $( \check{J}^c_{\lambda \lambda} \mathbf{a} \mathbf{a}^\top )$ is an approximation for $\check{J}^c_\mathbf{bb}$. Note that in order to compute the above approximation for $\check{J}^c_\mathbf{bb}$, we require only a single evaluation of $\underline{\lambda}_2^{\mathcal{L}}$ in~(\ref{eqn:hessian-lambda}). Considering the entire planning horizon, this results in only $T$ evaluations of $\underline{\lambda}_2^{\mathcal{L}}$ per iteration of belief-space iLQG. This is in contrast to using numerical differentiation which requires $O[\eta^2 n^4 T]$ evaluations as discussed in Section~\ref{subsec:bsilqg}. Thus, approximating $\check{J}^c_\mathbf{bb}$ with $( \check{J}^c_{\lambda \lambda} \mathbf{a} \mathbf{a}^\top )$ significantly reduces the computational load of the belief-space iLQG method.

\new{In summary, this section presented our distributed ADMM-based trajectory planning algorithm for connectivity maintenance. Note that the ADMM optimization step in~(\ref{eqn:admm-opt}) is non-convex and that our algorithm involves additional components such as the line search algorithm and the approximation of the Hessian. Thus, it is non-trivial to provide guarantees of whether our planning algorithm convergences to the optimal set of trajectories. However, as discussed in Section~\ref{subsec:admm-setup}, our planning algorithm ensures that the planned trajectories $\bar{U}$ satisfy the connectivity maintenance requirement from~(\ref{eqn:connectivity-maintenance}).}



%% file: simulations.tex
\label{sec:simulations}

\begin{figure}[b!]
    \centering
    \includegraphics[width=0.6\linewidth]{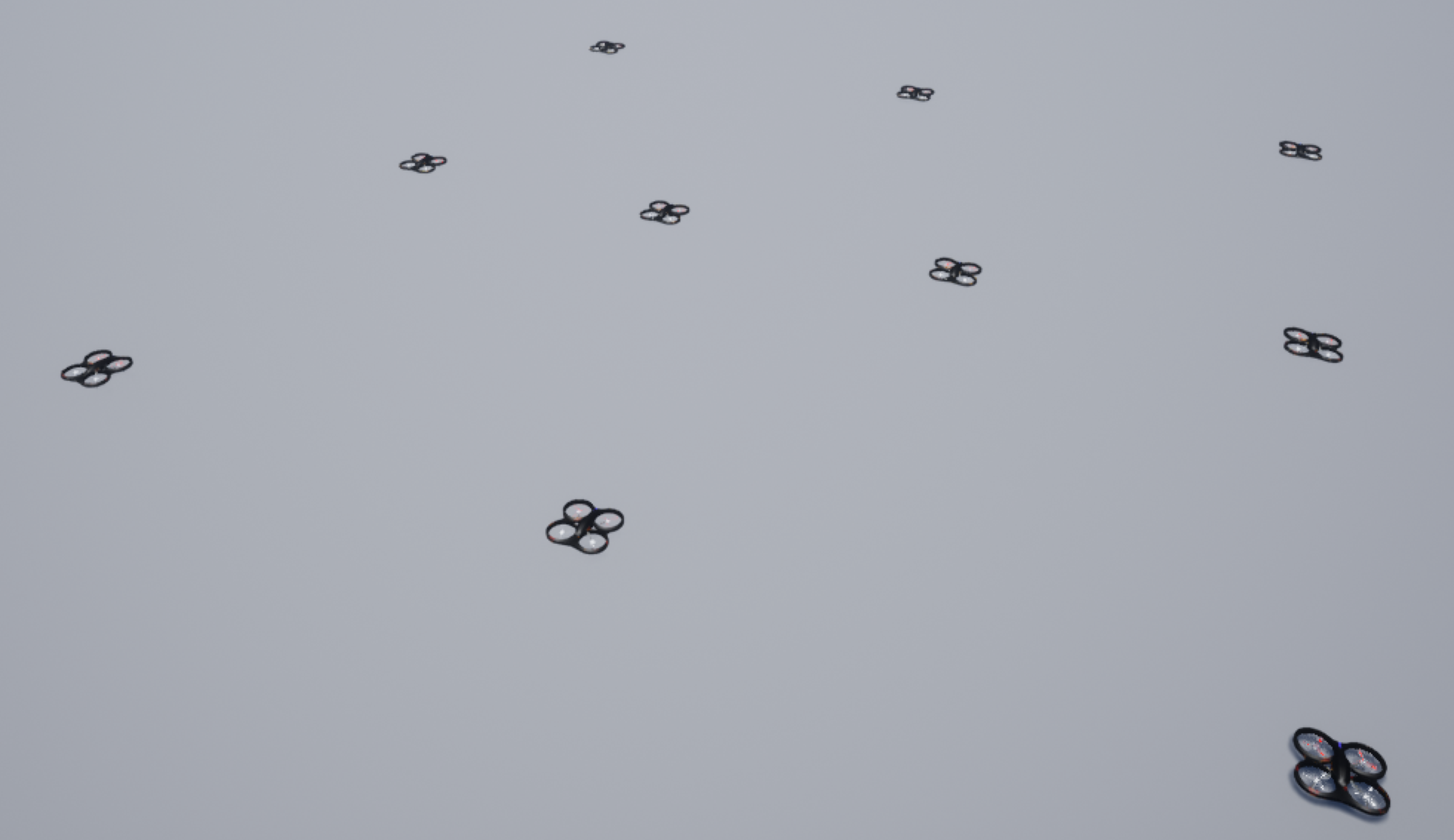}
    \caption{\new{Snapshot of the AirSim simulator~\cite{shah2018airsim} used to evaluate our trajectory planning algorithm on a high-fidelity simulator.}}
    \label{fig:airsim-setup}
    \vspace{-0.1cm}
\end{figure}

\new{In this section, we demonstrate the applicability of our trajectory planning algorithm for a multi-UAV mission under real-time constraints. We first describe details of the multi-UAV setup including the motion and sensing models. Next, we provide details of the simulated mission including the local tasks for the UAVs, the method used for generating initial trajectory guesses, and the values used for parameters within the planner. Finally, we discuss the performance of our planner across the simulated mission. Here we validate the connectivity maintenance of our algorithm by simulating $1000$ trajectory rollouts in MATLAB, where each rollout shows a possible realization of the multi-UAV system's trajectory under motion and sensing uncertainties. Additionally, we evaluate our planner on AirSim~\cite{shah2018airsim}, a high-fidelity simulator that represents UAV motion more realistically. Fig.~\ref{fig:airsim-setup} shows a snapshot for the multi-UAV setup in AirSim. All simulations in this section are performed on a \SI{2.80}{\GHz} Quad-core Intel\textsuperscript{TM} i7 machine.}

\new{
\vspace{-0.35cm}
\subsection{Multi-UAV system setup}
\label{subsec:sim-setup}}

For each UAV in the multi-UAV system, we consider a 2-dimensional (2D) double integrator model as the motion model. The UAV state vector contains the 2D position and velocity, i.e, $\mathbf{x}_{i,t} = \begin{bmatrix} \mathbf{p}_{i,t}^\top & \dot{\mathbf{p}}_{i,t}^\top \end{bmatrix}^\top$, and the input vector is the UAV accelerations. The motion model for the UAV can be written in the form of~(\ref{eqn:motion-model}) as:
\begin{equation}
    \label{eqn:sim-motion-model}
    \mathbf{x}_{i,t} = 
    \begingroup
    \setlength\arraycolsep{3pt}
    \begin{bmatrix} 1 & 0 & dt & 0 \\ 0 & 1 & 0 & dt \\ 0 & 0 & 1 & 0 \\ 0 & 0 & 0 & 1 \end{bmatrix} 
    \endgroup
    \mathbf{x}_{i,t-1} + 
    \begingroup
    \setlength\arraycolsep{1pt}
    \begin{bmatrix} \frac{dt^2}{2} & 0 \\ 0 & \frac{dt^2}{2} \\ dt & 0 \\ 0 & dt \end{bmatrix}
    \endgroup
    \mathbf{u}_{i,t-1} + \mathbf{w}_{i,t},
    \vspace{-0.15cm}
\end{equation}
\vspace{-0.1cm}
where $dt$ is the time-step between two time instants and
\begin{equation*}
    \mathbf{w}_{i,t} \sim \mathcal{N} \left[ \mathbf{0}, \SI{0.1}{\metre^{2} \second^{-3}}
    \begingroup
    \setlength\arraycolsep{1pt}
    \begin{bmatrix} \frac{dt^3}{3} & 0 & \frac{dt^2}{2} & 0\\ 0 & \frac{dt^3}{3} & 0 & \frac{dt^2}{2} \\ \frac{dt^2}{2} & 0 & dt & 0 \\ 0 & \frac{dt^2}{2} & 0 & dt \end{bmatrix}
    \endgroup
    \right].
\end{equation*}


\new{As discussed in Section~\ref{subsec:ekf-belief-dynamics}, using a linear model to represent the UAV motion can be restrictive. However, a double-integrator model as in (\ref{eqn:sim-motion-model}) has been previously used to represent UAVs~\cite{park2019distributed,hou2015distributed} and is more realistic than the single-integrator model used in the majority of related work.} Note that it is important to choose an appropriately small time-step $dt$ such that the system connectivity along the discretized trajectory represents the system connectivity in continuous time. For our simulations we choose $dt = \SI{0.2}{\second}$ which also reflects the rate of \new{common localization measurements from GNSS or camera.} Additionally, we set a maximum limit of \SI{5}{\metre \per \square \second} on the magnitude of input accelerations. For the sensing model in~(\ref{eqn:measurement-model}), we consider position measurements:
\begin{equation}
    \label{eqn:sim-sensing-model}
    \mathbf{z}_{i,t} = \begin{bmatrix} 1 & 0 & 0 & 0 \\ 0 & 1 & 0 & 0 \end{bmatrix} \mathbf{x}_{i,t} + \mathbf{v}_{i,t},
\end{equation}
where $\mathbf{v}_{i,t} \sim \mathcal{N} \left[ \mathbf{0}, \texttt{diag}(\SI{1}{\square\metre}, \SI{1}{\square\metre}) \right].$

\new{
\vspace{-0.25cm}
\subsection{Simulated mission details}
\label{subsec:simulated-mission}
}

\begin{figure}[t!]
    \centering
    \includegraphics[width=0.9\linewidth]{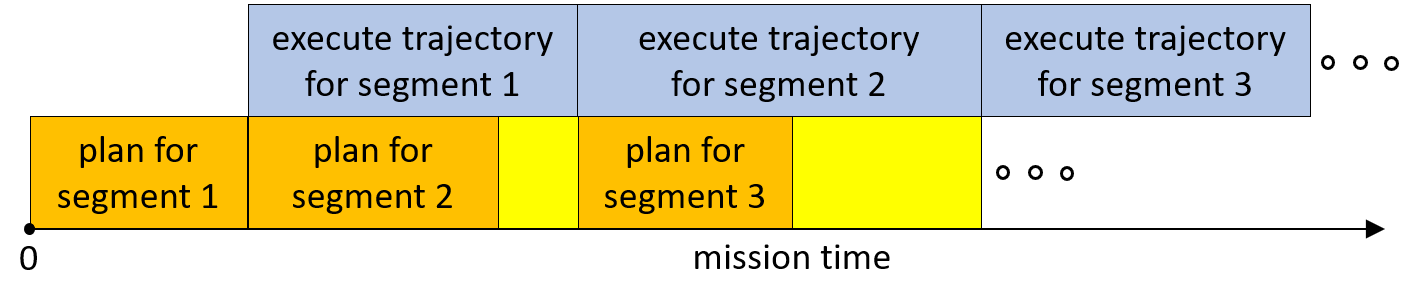}
    \caption{\new{Order of trajectory planning (orange) and execution (blue) for the simulated real-time mission with multiple segments. The multi-UAV system plans trajectories for the upcoming segment while executing trajectories for the current segment. We allow a maximum planning time for each segment since the remaining time (yellow) could be required for other purposes such as analyzing sensor data.}}
    \label{fig:online-mission-timeline}
    \vspace{-0.5cm}
\end{figure}

We consider the mission to be comprised of multiple segments and allow a maximum planning time \new{(stopping criterion for the planner)} for each segment. This resembles real-time applications such as exploration, coverage, or formation control, where only a limited amount of computation time is available for planning trajectories while the remaining time is required for other purposes such as analyzing sensor data or decision making. \new{Fig.~\ref{fig:online-mission-timeline} shows how our trajectory planning algorithm is used across multiple segments of the mission. The trajectories for the first segment are planned while the system remains stationary. For the remaining segments, the trajectories for the upcoming segment are planned while executing the planned trajectory for the current segment.}

We consider the multi-UAV system to consist of ten UAVs: six \textit{primary} and four \textit{bridge}. \new{The goal of the \textit{primary} robots is to reach desired positions, whereas the \textit{bridge} robots focus on maintaining connectivity in the system. Here we assume that the desired positions are available from a high-level planning strategy (such as exploration or formation control), and pick them randomly for our simulation. Additionally, we assume that each UAV operates at a different altitude, similar to~\cite{park2019distributed}. We make this assumption in order to alleviate inter-UAV collision constraints and focus on the connectivity maintenance of the system. For each UAV, we consider the local task of reaching a desired position along with minimizing the control input effort. Thus, $J_{i,t}$ $\forall$ $t \in [0,T]$ in Section~\ref{subsec:trajecory-planning} are set as:} 
\begin{align}
    \label{eqn:sim-cost-input}
    J_{i,t} & = \mathbf{u}_{i,t}^\top W^{\mathbf{u}}_i \mathbf{u}_{i,t}, \ \ \forall t \in [0,T-1] \\
    \label{eqn:sim-cost-position}
    J_{i,T} & = ( \hat{\mathbf{x}}_{i,T} - \mathbf{x}_{i,\text{des}} )^\top W^{\mathbf{x}}_i ( \hat{\mathbf{x}}_{i,T} - \mathbf{x}_{i,\text{des}} ),
\end{align}
where $\mathbf{x}_{i,\text{des}} = \begin{bmatrix} \mathbf{p}_{i,\text{des}}^\top & \dot{\mathbf{p}}_{i,\text{des}}^\top \end{bmatrix}^\top$ contains the desired position $\mathbf{p}_{i,\text{des}}$ and desired velocity $\dot{\mathbf{p}}_{i,\text{des}}$ for the UAV, and $W^{\mathbf{x}}_i$ and $W^{\mathbf{u}}_i$ are used to set the relative importance of the different costs. $\mathbf{p}_{i,\text{des}}$ are the randomly picked desired positions and $\dot{\mathbf{p}}_{i,\text{des}}$ is set to $\mathbf{0}$. We specify an initial position $\mathbf{p}_{i,\text{init}}$ for each UAV and set their initial velocities to be zeros. The initial state estimation covariance is set as $P_{i,\text{init}} = \texttt{diag}(\SI{0.1}{\metre^2}, \SI{0.1}{\metre^2}, \SI{0.001}{\metre^2\second^{-2}}, \SI{0.001}{\metre^2\second^{-2}})$.

For the initial trajectory guess, we require a computationally inexpensive method of generating a trajectory based on the local tasks for each UAV. For example, sampling-based planners such as rapidly-exploring random trees (RRTs) can be used in order to quickly plan trajectories around obstacles~\cite{van2012motion,goretkin2013optimal}; in a multiple target tracking application, each UAV can be randomly assigned to track a separate target~\cite{park2019distributed}. In our algorithm, we use a linear-quadratic-regulator (LQR) to obtain an initial trajectory guess for each \textit{primary} UAV from $\mathbf{x}_{i,\text{init}}$ to $\mathbf{x}_{i,\text{des}}$. For \textit{bridge} UAVs, we simply set the initial trajectory guess to be hovering at the initial position. If $\underline{\lambda}_2^{\mathcal{L}_t}$ is not maintained above $\epsilon$ for the resulting trajectory guess, we consider new desired states (only for the initial trajectory guess and not for the rest of the planner) midway between $\mathbf{x}_{i,\text{init}}$ and the $\mathbf{x}_{i,\text{des}}$ for all \textit{primary} UAVs and repeat the process.

\new{We consider the mission to consist of six segments. For each segment, we} set a maximum planning time of \SI{25}{\second} and a trajectory duration of \SI{50}{\second} ($T = \num{250}$). \new{Additionally, since our planner uses a centralized communication architecture, we account for a delay of \SI{0.2}{\second} (by simply pausing the code execution) in each ADMM iteration of our planning algorithm.} We consider a communication range of $\Delta = \SI{40}{\metre}$ and set the parameter $\Delta_0 = \SI{35}{\metre}$ in~(\ref{eqn:conservative-edge-weight}). For the connectivity maintenance requirement in~(\ref{eqn:connectivity-maintenance}), we specify the lower algebraic connectivity limit $\epsilon = \num{0.1}$ and the corresponding probability value $\delta = \num{0.003}$ to reflect a $3\sigma$ confidence level. In~(\ref{eqn:connectivity-cost}), we set the parameter for the magnitude of the connectivity cost as $k_c = \num{0.001}$. We set $\eta = \num{2}$ for the number of elements in subsets $\mathcal{V}$ obtained in the trajectory planner. Finally, for the line search algorithm used to update the ADMM consensus variable in~(\ref{eqn:admm-line-search}), we set $\gamma = \num{0.8}$.

\vspace{-0.3cm}
\subsection{Planning results under real-time constraints}
\label{subsec:online-planning}

\begin{figure*}[]
  \centering
  \subfloat{%
        \includegraphics[width=0.7\linewidth,trim={0cm 1cm 0cm 0},clip]{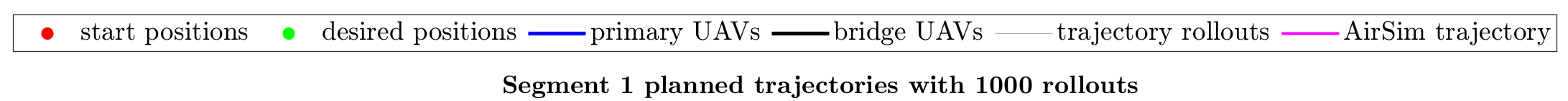}} \addtocounter{subfigure}{-1}
        \hfill \\
        \vspace{-0.35cm}
  \hspace{1.5cm}
  \subfloat[]{%
        \includegraphics[width=0.23\linewidth,trim={3.5cm 0 2.5cm 0},clip]{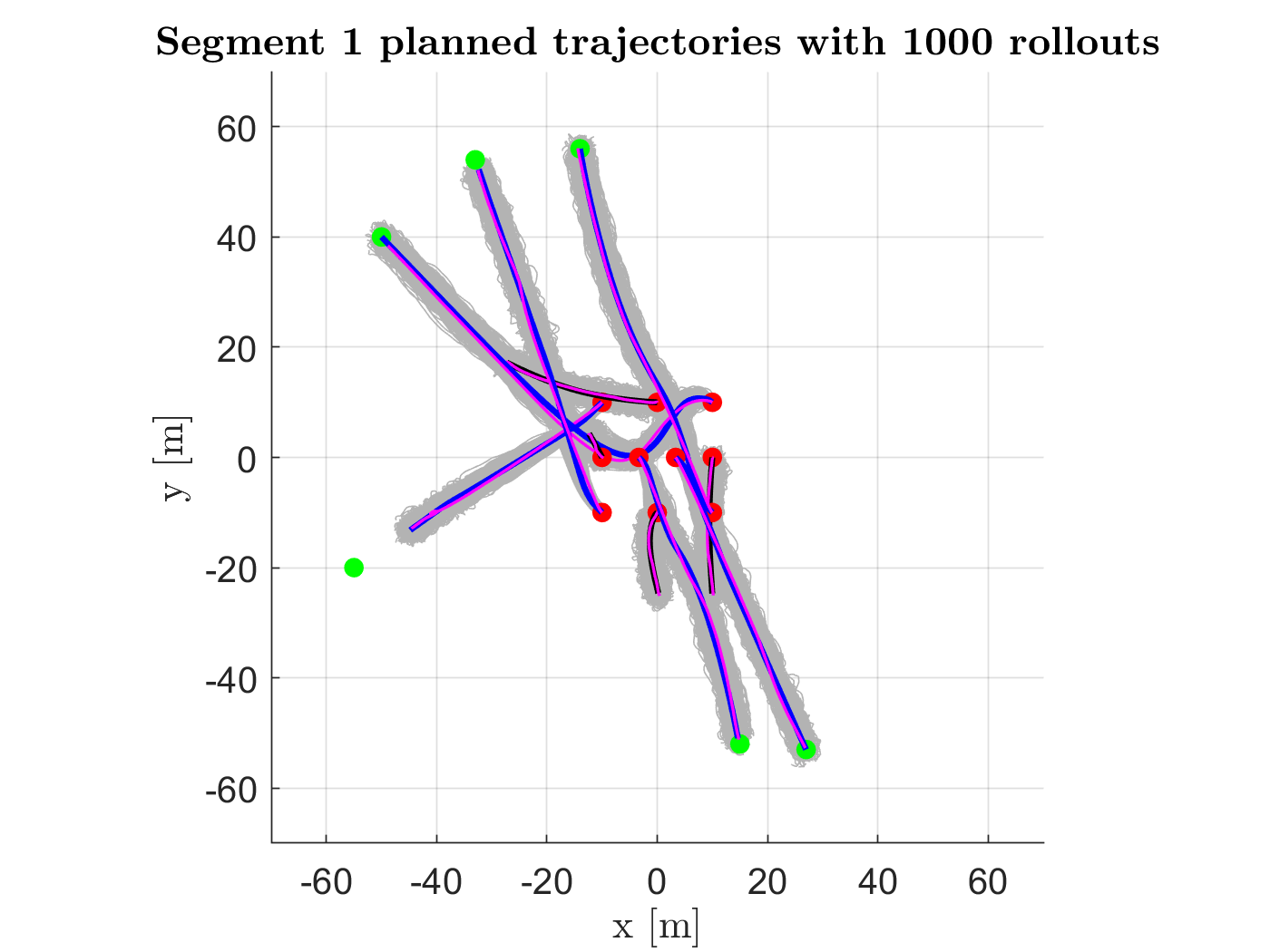}}
        \hfill
  \subfloat[]{%
        \includegraphics[width=0.23\linewidth,trim={3.5cm 0 2.5cm 0},clip]{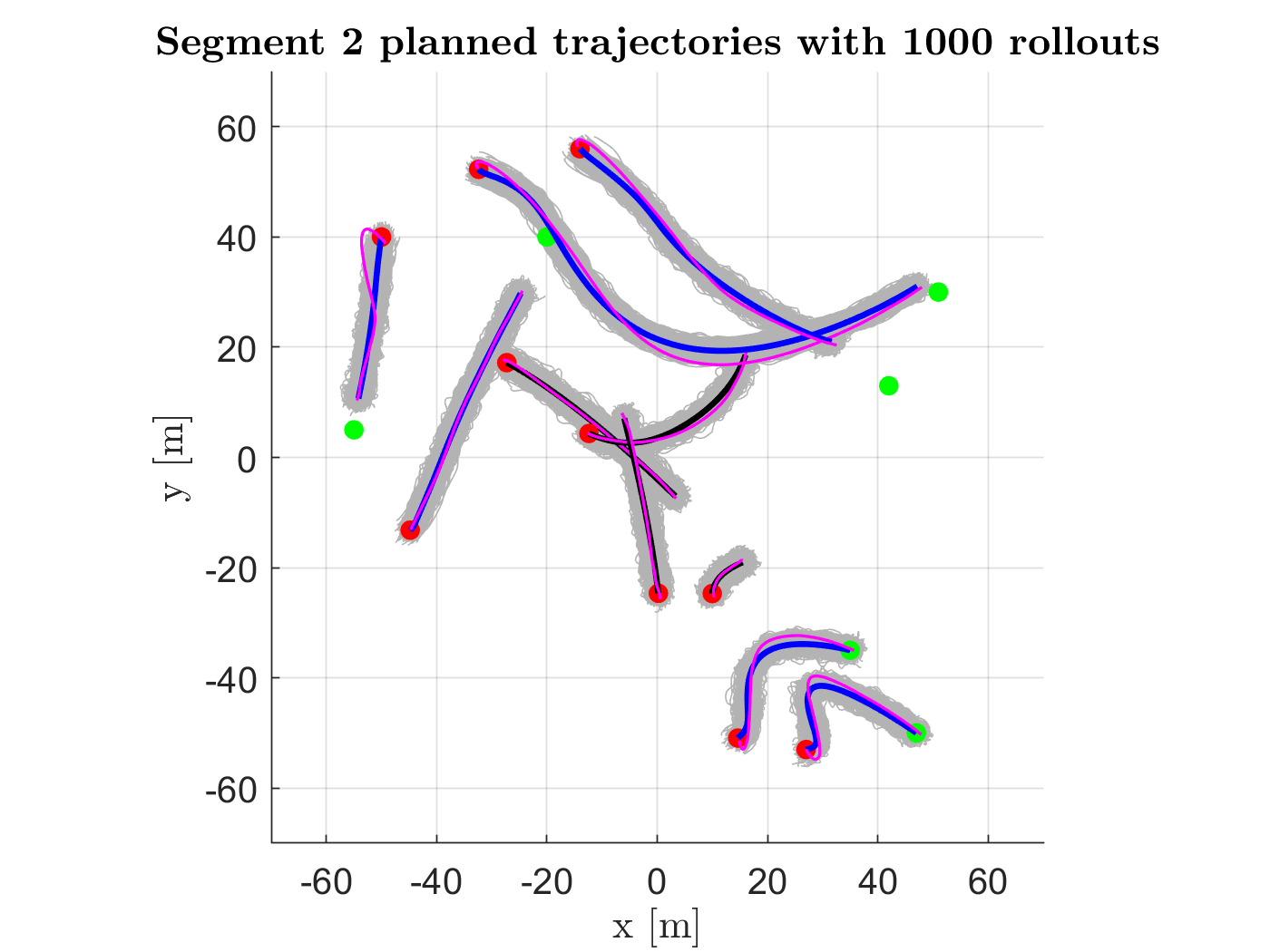}}
        \hfill
  \subfloat[]{%
        \includegraphics[width=0.23\linewidth,trim={3.5cm 0 2.5cm 0},clip]{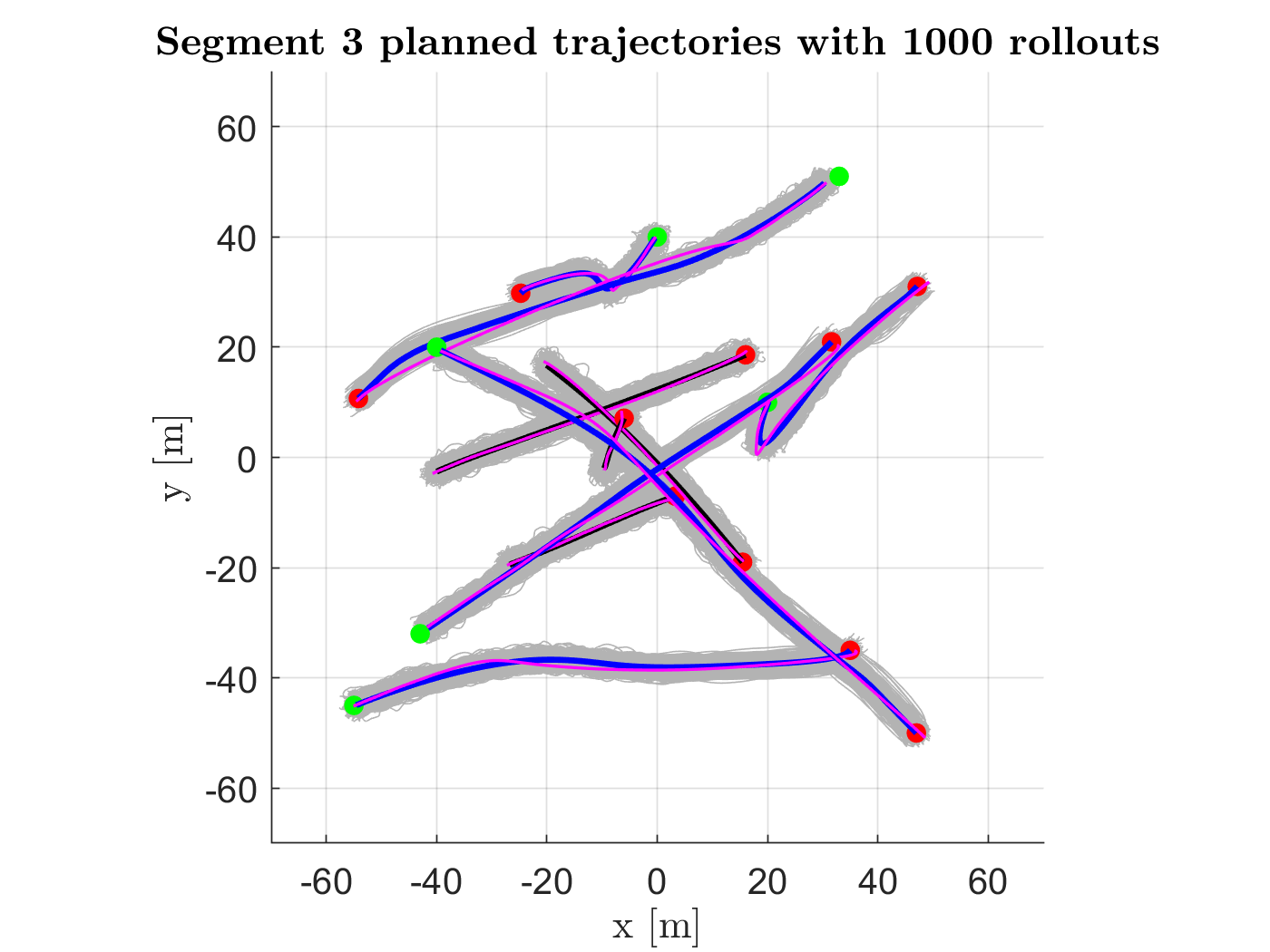}}
        \hfill \hspace{1.5cm} \\
        \vspace{-0.35cm}
  \hspace{1.5cm}
  \subfloat[]{%
        \includegraphics[width=0.23\linewidth,trim={3.5cm 0 2.5cm 0},clip]{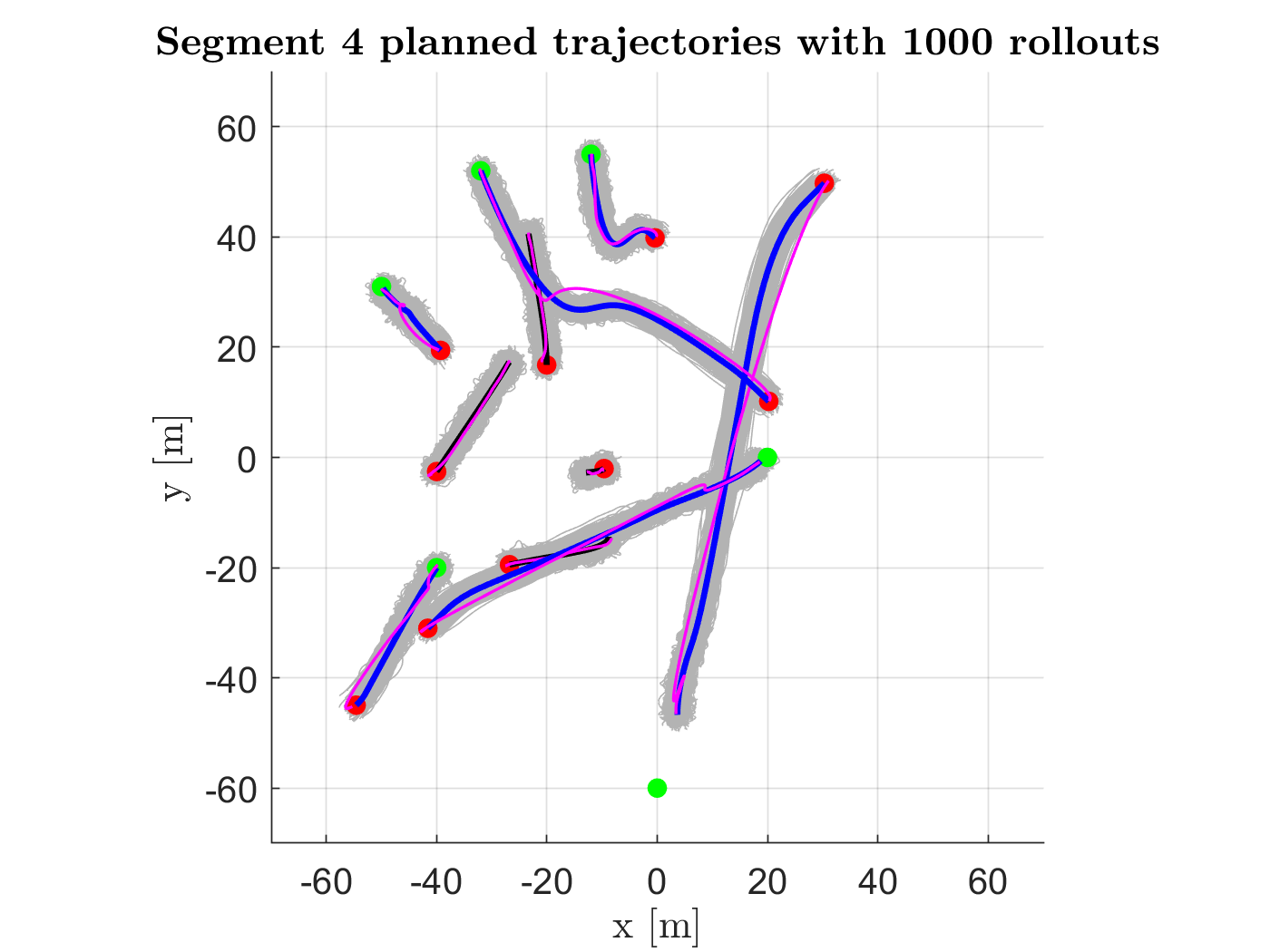}}
        \hfill
  \subfloat[]{%
        \includegraphics[width=0.23\linewidth,trim={3.5cm 0 2.5cm 0},clip]{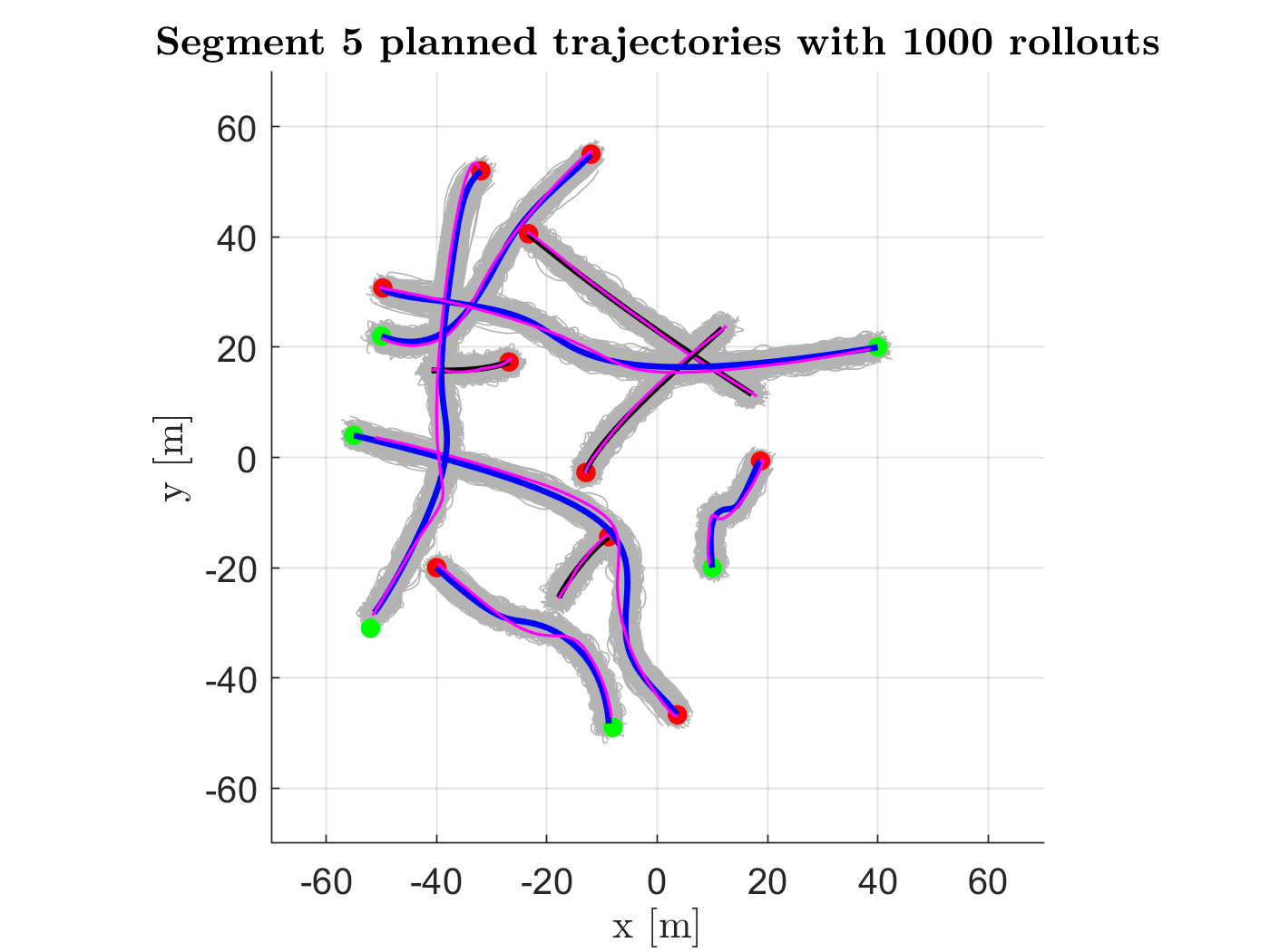}}
        \hfill
  \subfloat[]{%
        \includegraphics[width=0.23\linewidth,trim={3.5cm 0 2.5cm 0},clip]{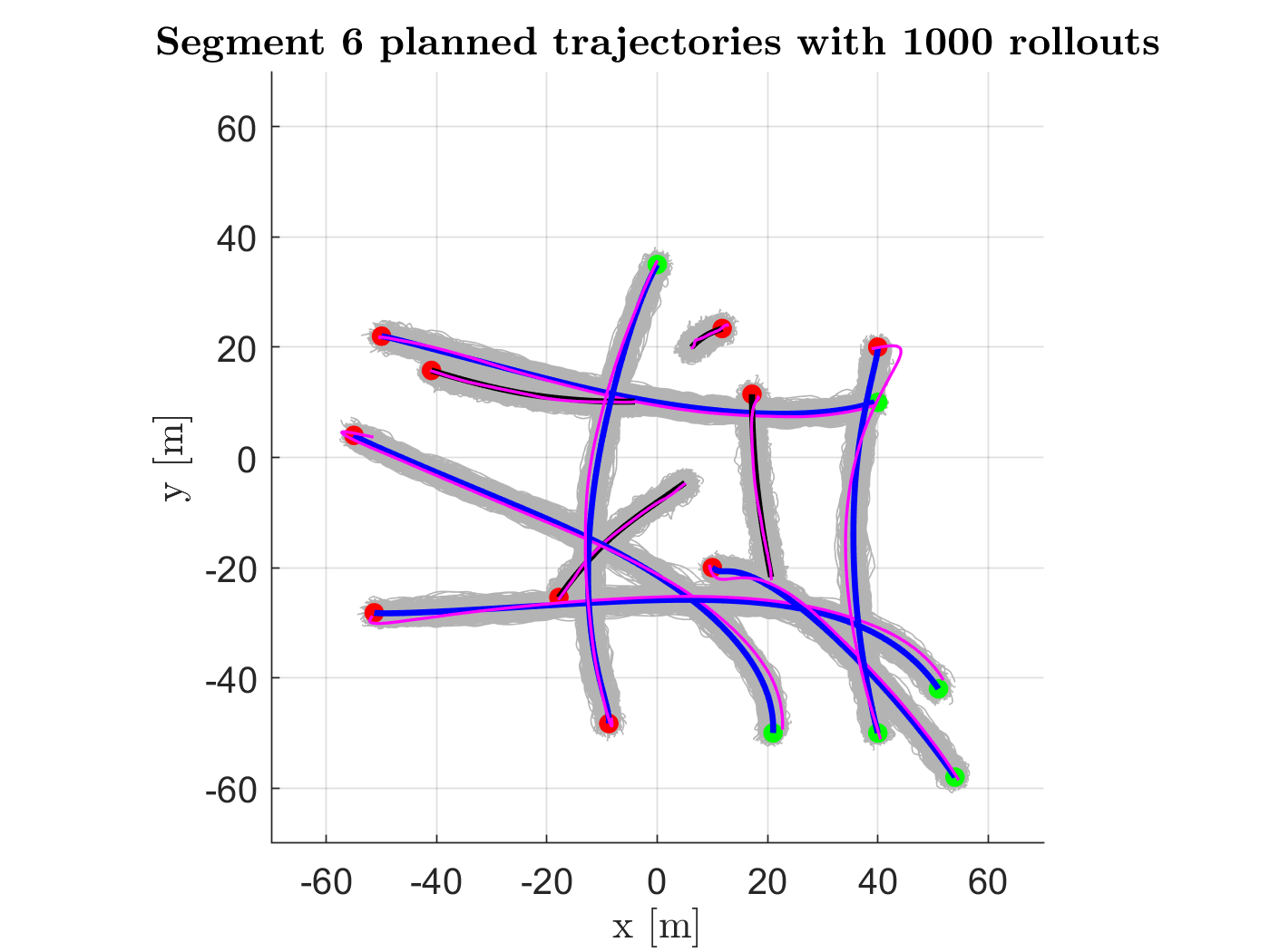}}
        \hfill \hspace{1.5cm} \\
        \vspace{-0.1cm}
  \subfloat[]{%
        \includegraphics[width=0.16\linewidth,trim={0 0 1cm 0},clip]{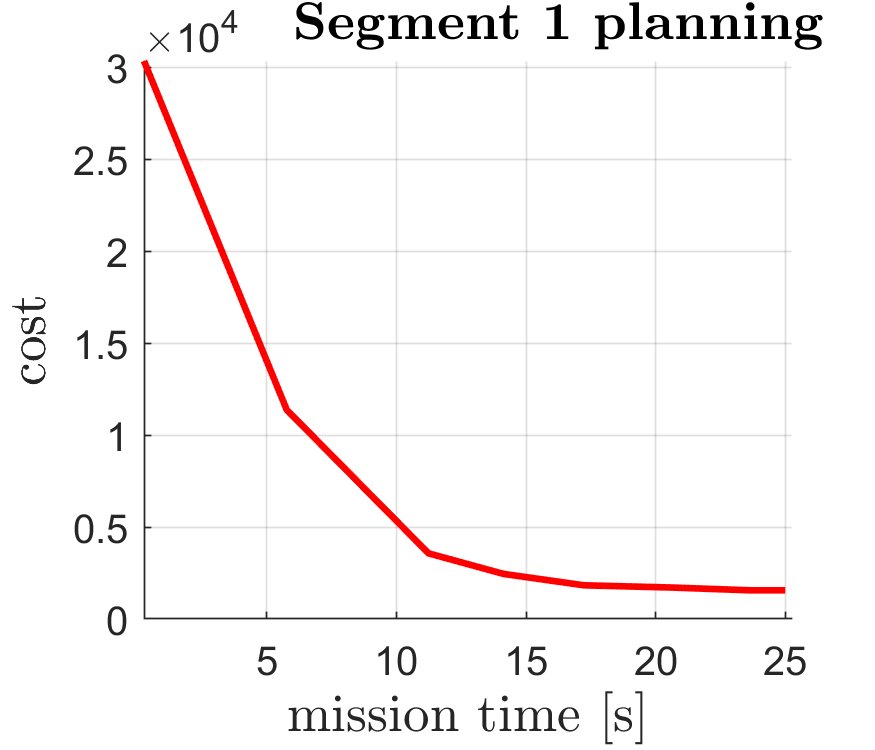}}
        \hfill
  \subfloat[]{%
        \includegraphics[width=0.16\linewidth,trim={0 0 1cm 0},clip]{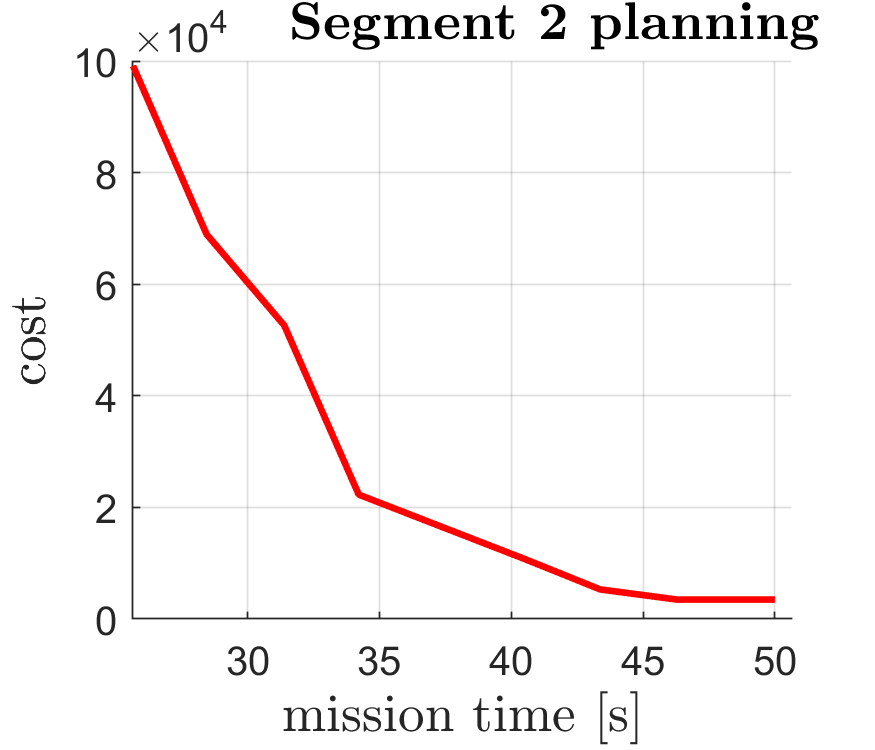}}
        \hfill
  \subfloat[]{%
        \includegraphics[width=0.16\linewidth,trim={0 0 1cm 0},clip]{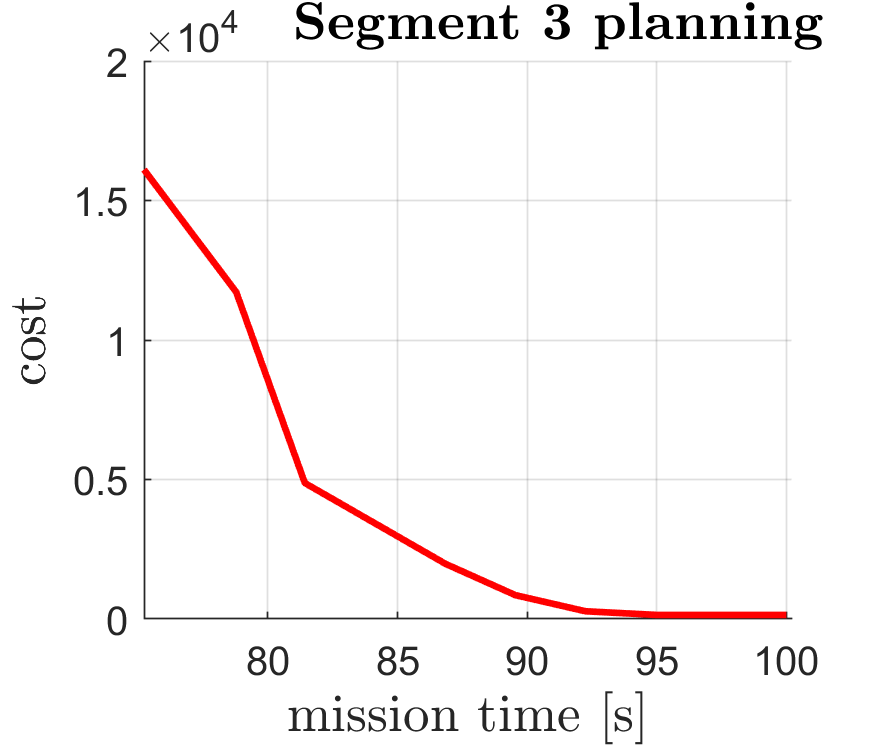}}
        \hfill
  \subfloat[]{%
        \includegraphics[width=0.16\linewidth,trim={0 0 1cm 0},clip]{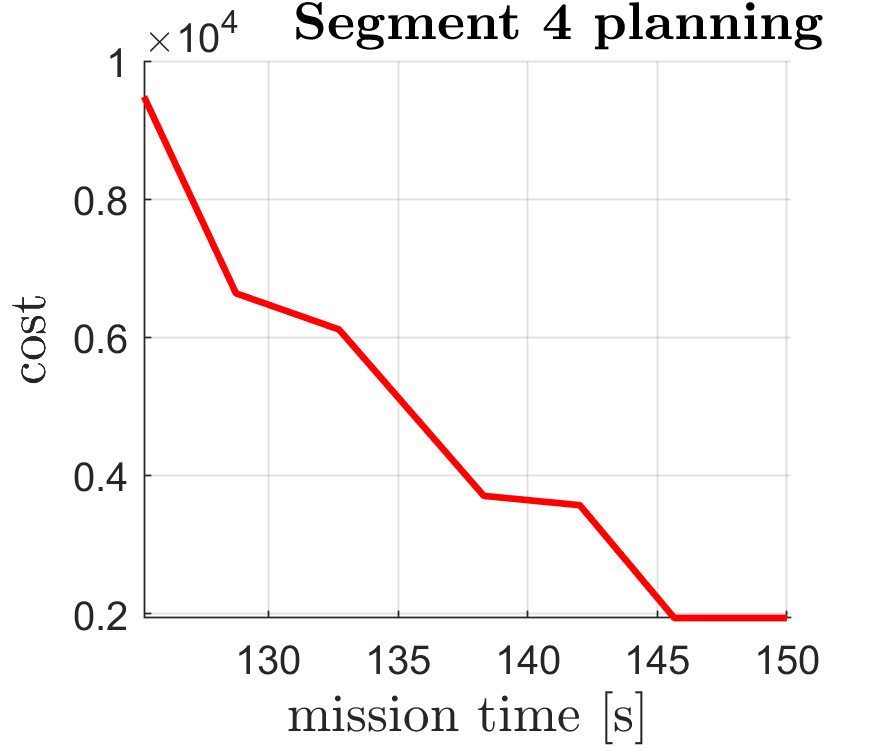}}
        \hfill
  \subfloat[]{%
        \includegraphics[width=0.16\linewidth,trim={0 0 1cm 0},clip]{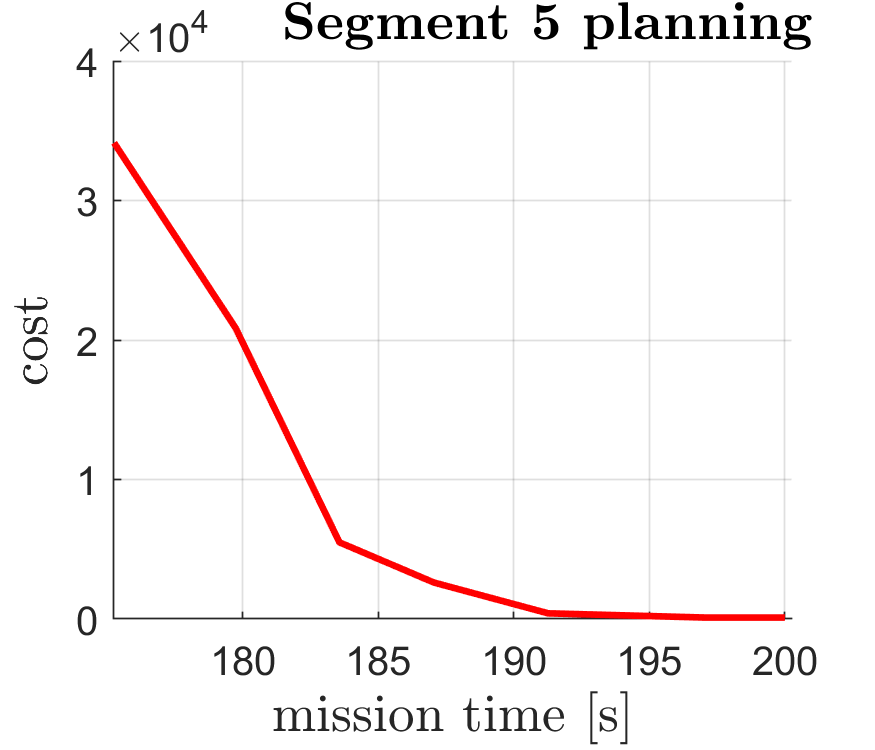}}
        \hfill
  \subfloat[]{%
        \includegraphics[width=0.16\linewidth,trim={0 0 1cm 0},clip]{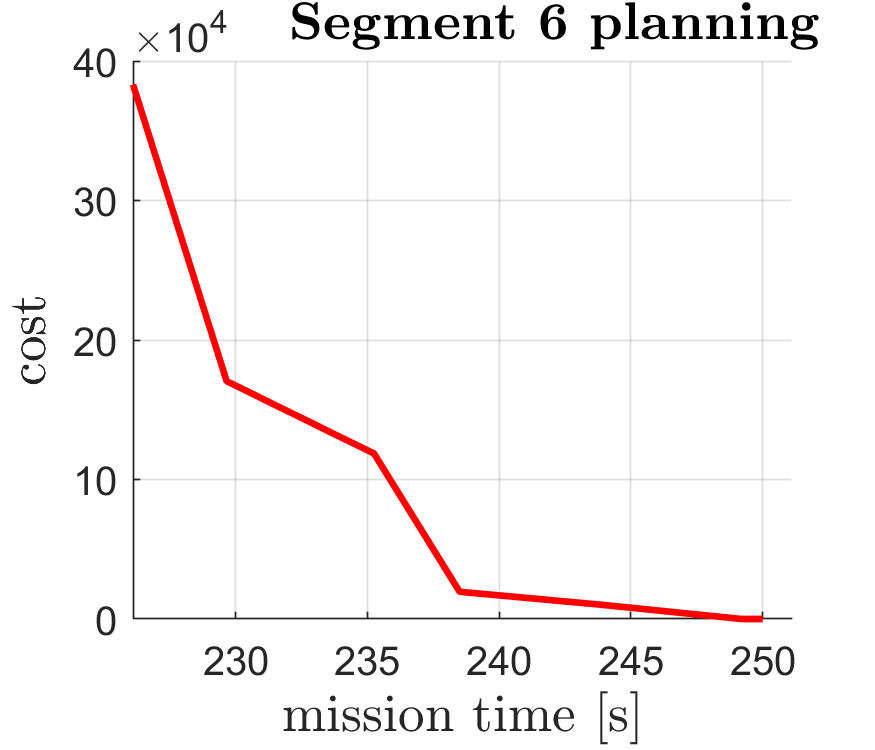}}
        \hfill \\
        \vspace{-0.01cm}
  \hspace{2.7cm}
  \subfloat[]{%
        \includegraphics[width=0.7\linewidth]{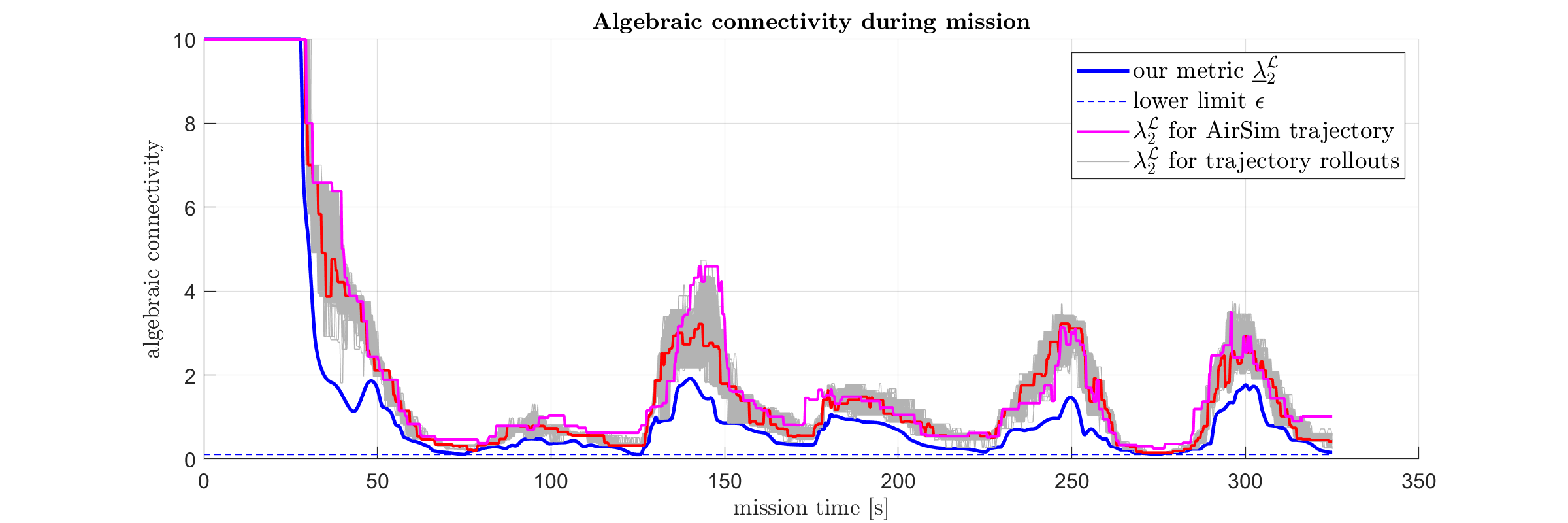}}
        \hfill
        \vspace{-0.2cm}
  \caption{Trajectory planning and connectivity maintenance for the simulated mission with six \textit{primary} and four \textit{bridge} UAVs. (a)-(f) The final planned trajectories, $1000$ trajectory rollouts and AirSim trajectory for each mission segment. (g)-(l) Convergence of the cost in the transformed optimization problem for a maximum planning time of \SI{25}{\second}. (m) Connectivity maintenance performance throughout the mission. Only one (red) of the $1000$ trajectory rollouts results in system algebraic connectivity ${\lambda}_2^{\mathcal{L}}$ less than $\underline{\lambda}_2^{\mathcal{L}}$, whereas none drop below the specified lower limit $\epsilon = 0.1$.}
  \label{fig:mission-4}
  \vspace{-0.5cm}
\end{figure*}


\new{Fig.~\ref{fig:mission-4} shows the results of our planner for the simulated real-time mission. In Figs.~\ref{fig:mission-4}(a)-(f) we show the planned trajectories for the six mission segments (blue for \textit{primary} UAVs and black for \textit{bridge} UAVs), along with $1000$ rollouts (gray) showing possible trajectory realizations of the system. Additionally, we show the trajectories of the system simulated in AirSim (magenta). For all the segments we observe that the planner attempts to drive the \textit{primary} UAVs to their desired positions while rearranging the \textit{bridge} UAVs for maintaining connectivity. For cases when the desired positions might lead to a loss of connectivity (such as the leftmost desired position in Fig.~\ref{fig:mission-4}(a)), our planner attempts to bring the \textit{primary} UAVs as close as possible while maintaining connectivity. Also, note that the planned trajectories closely match the UAV motion from AirSim, thus validating the use of a double-integrator motion model in~(\ref{eqn:sim-motion-model}).}

\new{Figs.~\ref{fig:mission-4}(g)-(l) show the convergence of the cost from~(\ref{eqn:transformed-opt-problem}) for the six mission segments. We observe that our planner is able to find lower cost trajectories within the \SI{25}{\second} planning time for various desired system configurations. In Fig.~\ref{fig:mission-4}(m) we analyze the connectivity maintenance throughout the mission. Our planner maintains the algebraic connectivity of our weighted graph $\underline{\lambda}_2^{\mathcal{L}}$ (blue) above the specified lower limit $\epsilon = 0.1$ (blue-dashed). In order to validate that the connectivity maintenance requirement from~(\ref{eqn:connectivity-maintenance}) is satisfied, we plot the algebraic connectivity of the $1000$ trajectory rollouts (gray), which are obtained using~(\ref{eqn:binary-edge-weight}). Note that since the edge weights in~(\ref{eqn:binary-edge-weight}) are binary, the corresponding algebraic connectivity contains jumps when an edge weight changes from $0$ to $1$, or from $1$ to $0$. We observe that the algebraic connectivity for only one rollout (red) drops below $\underline{\lambda}_2^{\mathcal{L}}$, while none drop below $\epsilon$. This validates that our planner satisfies the connectivity maintenance requirement from~(\ref{eqn:connectivity-maintenance}). Additionally, we plot the algebraic connectivity of the multi-UAV system simulated in AirSim (magenta) and observe that it also remains above the lower limit $\epsilon$.}

\vspace{-0.1cm}

%% file: conclusions.tex
\label{sec:conclusions}

\new{We have presented a trajectory planning algorithm for global connectivity maintenance of multi-robot systems that addresses two limitations in related work: it accounts for robot motion and sensing uncertainties, and it considers general linear robot motion models which do not necessarily have an instantaneously changeable direction of motion. For connectivity maintenance, we first define a weighted undirected graph to represent connectivity of a system with uncertain robot positions.} The algebraic connectivity of this graph is then used to define a transformed trajectory planning problem which is solved by a distributed ADMM setup. \new{We present an approach to reduce the computational load of the ADMM optimization step by approximating the required Hessian matrices. Finally, we evaluate the planner under real-time constraints on a simulated multi-UAV mission with multiple segments. Our planner plans trajectories attempting the complete the local UAV tasks while satisfying the connectivity maintenance requirement.}

While we have demonstrated the utility of our planner in addressing the aforementioned limitations in related work, multiple future directions of work exist. First, a natural extension includes exploring decentralized architectures in order to improve the \new{scalability with respect to the communication load. Second, we plan to extend our algorithm to more general robot motion models such as feedback linearizable or differentially flat systems. We also plan to evaluate our planner for nonlinear systems using an approximation of the state uncertainty provided by the EKF.} Third, it is desirable to include additional realistic constraints for the multi-robot system such as line-of-sight communication and collision avoidance. Finally, we also plan to test the planner on a real-world hardware multi-robot platform.

\vspace{-0.2cm}

%% file: bare_jrnl.bbl
\begin{thebibliography}{10}
\providecommand{\url}[1]{#1}
\csname url@samestyle\endcsname
\providecommand{\newblock}{\relax}
\providecommand{\bibinfo}[2]{#2}
\providecommand{\BIBentrySTDinterwordspacing}{\spaceskip=0pt\relax}
\providecommand{\BIBentryALTinterwordstretchfactor}{4}
\providecommand{\BIBentryALTinterwordspacing}{\spaceskip=\fontdimen2\font plus
\BIBentryALTinterwordstretchfactor\fontdimen3\font minus
  \fontdimen4\font\relax}
\providecommand{\BIBforeignlanguage}[2]{{%
\expandafter\ifx\csname l@#1\endcsname\relax
\typeout{** WARNING: IEEEtran.bst: No hyphenation pattern has been}%
\typeout{** loaded for the language `#1'. Using the pattern for}%
\typeout{** the default language instead.}%
\else
\language=\csname l@#1\endcsname
\fi
#2}}
\providecommand{\BIBdecl}{\relax}
\BIBdecl

\bibitem{rizk2019cooperative}
Y.~Rizk, M.~Awad, and E.~W. Tunstel, ``{Cooperative Heterogeneous Multi-robot
  Systems: A Survey},'' \emph{ACM Computing Surveys (CSUR)}, vol.~52, no.~2,
  pp. 1--31, 2019.

\bibitem{alanwar2019distributed}
A.~Alanwar, H.~Said, and M.~Althoff, ``{Distributed Secure State Estimation
  Using Diffusion Kalman Filters and Reachability Analysis},'' in \emph{2019
  IEEE 58th Conference on Decision and Control (CDC)}.\hskip 1em plus 0.5em
  minus 0.4em\relax IEEE, 2019, pp. 4133--4139.

\bibitem{park2018robust}
H.~Park and S.~Hutchinson, ``{Robust Rendezvous for Multi-robot System with
  Random Node Failures: An Optimization Approach},'' \emph{Autonomous Robots},
  vol.~42, no.~8, pp. 1807--1818, 2018.

\bibitem{khateri2019comparison}
K.~Khateri, M.~Pourgholi, M.~Montazeri, and L.~Sabattini, ``{A Comparison
  Between Decentralized Local and Global Methods for Connectivity Maintenance
  of Multi-robot Networks},'' \emph{IEEE Robotics and Automation Letters},
  vol.~4, no.~2, pp. 633--640, 2019.

\bibitem{de2006decentralized}
M.~C. De~Gennaro and A.~Jadbabaie, ``{Decentralized Control of Connectivity for
  Multi-agent Systems},'' in \emph{Proceedings of the 45th IEEE Conference on
  Decision and Control}.\hskip 1em plus 0.5em minus 0.4em\relax IEEE, 2006, pp.
  3628--3633.

\bibitem{yang2010decentralized}
P.~Yang, R.~A. Freeman, G.~J. Gordon, K.~M. Lynch, S.~S. Srinivasa, and
  R.~Sukthankar, ``{Decentralized Estimation and Control of Graph Connectivity
  for Mobile Sensor Networks},'' \emph{Automatica}, vol.~46, no.~2, pp.
  390--396, 2010.

\bibitem{Sabattini2013}
\BIBentryALTinterwordspacing
L.~Sabattini, N.~Chopra, and C.~Secchi, ``{Decentralized Connectivity
  Maintenance for Cooperative Control of Mobile Robotic Systems},'' \emph{The
  International Journal of Robotics Research}, vol.~32, no.~12, pp. 1411--1423,
  oct 2013. [Online]. Available:
  \url{http://journals.sagepub.com/doi/10.1177/0278364913499085}
\BIBentrySTDinterwordspacing

\bibitem{Sabattini2013a}
L.~Sabattini, C.~Secchi, N.~Chopra, and A.~Gasparri, ``{Distributed Control of
  Multirobot Systems with Global Connectivity Maintenance},'' \emph{IEEE
  Transactions on Robotics}, vol.~29, no.~5, pp. 1326--1332, 2013.

\bibitem{robuffo2013passivity}
P.~Robuffo~Giordano, A.~Franchi, C.~Secchi, and H.~H. B{\"u}lthoff, ``{A
  Passivity-based Decentralized Strategy for Generalized Connectivity
  Maintenance},'' \emph{The International Journal of Robotics Research},
  vol.~32, no.~3, pp. 299--323, 2013.

\bibitem{gasparri2017bounded}
A.~Gasparri, L.~Sabattini, and G.~Ulivi, ``{Bounded Control Law for Global
  Connectivity Maintenance in Cooperative Multirobot Systems},'' \emph{IEEE
  Transactions on Robotics}, vol.~33, no.~3, pp. 700--717, 2017.

\bibitem{capelli2020connectivity}
B.~Capelli and L.~Sabattini, ``{Connectivity Maintenance: Global and Optimized
  approach through Control Barrier Functions},'' \emph{Proceedings - IEEE
  International Conference on Robotics and Automation}, pp. 5590--5596, may
  2020.

\bibitem{Bhattacharya2010}
S.~Bhattacharya and T.~Başar, ``{Graph-theoretic Approach for Connectivity
  Maintenance in Mobile Networks in the Presence of a Jammer},'' in
  \emph{Proceedings of the IEEE Conference on Decision and Control}, 2010, pp.
  3560--3565.

\bibitem{thrun2005probabilistic}
S.~Thrun, W.~Burgard, and D.~Fox, ``{Probabilistic Robotics (Intelligent
  Robotics and Autonomous Agents series), ser. Intelligent Robotics and
  Autonomous Agents},'' 2005.

\bibitem{mclain2014implementing}
M.~Owen, R.~Beard, and T.~McLain, ``{Implementing Dubins Airplane Paths on
  Fixed-wing UAVs},'' \emph{Contributed chapter to the Handbook of Unmanned
  Aerial Vehicles}, pp. 1677--1701, 2014.

\bibitem{Boyd2010}
S.~Boyd, N.~Parikh, E.~Chu, J.~Eckstein, S.~Boyd, N.~Parikh, E.~Chu,
  B.~Peleato, and J.~Eckstein, ``{Distributed Optimization and Statistical
  Learning via the Alternating Direction Method of Multipliers},''
  \emph{Foundations and Trends R in Machine Learning}, vol.~3, no.~1, pp.
  1--122, 2010.

\bibitem{Luo}
W.~Luo, S.~Yi, and K.~Sycara, ``{Behavior Mixing with Minimum Global and
  Subgroup Connectivity Maintenance for Large-Scale Multi-Robot Systems},''
  \emph{Proceedings - IEEE International Conference on Robotics and
  Automation}, pp. 9845--9851, May 2020.

\bibitem{Scherer}
J.~Scherer and B.~Rinner, ``{Multi-Robot Persistent Surveillance with
  Connectivity Constraints},'' \emph{IEEE Access}, vol.~8, pp.
  15\,093--15\,109, 2020.

\bibitem{luo2019minimum}
W.~Luo and K.~Sycara, ``{Minimum \textit{k}-connectivity Maintenance for Robust
  Multi-Robot Systems},'' in \emph{2019 IEEE/RSJ International Conference on
  Intelligent Robots and Systems (IROS)}.\hskip 1em plus 0.5em minus
  0.4em\relax IEEE, 2019, pp. 7370--7377.

\bibitem{van2012motion}
J.~Van Den~Berg, S.~Patil, and R.~Alterovitz, ``{Motion Planning under
  Uncertainty using Iterative Local Optimization in Belief Space},'' \emph{The
  International Journal of Robotics Research}, vol.~31, no.~11, pp. 1263--1278,
  2012.

\bibitem{hou2015distributed}
Z.~Hou and I.~Fantoni, ``{Distributed Leader-Follower Formation Control for
  Multiple Quadrotors With Weighted Topology},'' in \emph{2015 10th System of
  Systems Engineering Conference (SoSE)}.\hskip 1em plus 0.5em minus
  0.4em\relax IEEE, 2015, pp. 256--261.

\bibitem{park2019distributed}
S.-S. Park, Y.~Min, J.-S. Ha, D.-H. Cho, and H.-L. Choi, ``{A Distributed ADMM
  Approach to Non-Myopic Path Planning for Multi-Target Tracking},'' \emph{IEEE
  Access}, vol.~7, pp. 163\,589--163\,603, 2019.

\bibitem{grone1990laplacian}
R.~Grone, R.~Merris, and V.~S. Sunder, ``{The Laplacian Spectrum of a Graph},''
  \emph{SIAM Journal on matrix analysis and applications}, vol.~11, no.~2, pp.
  218--238, 1990.

\bibitem{hoover1984algorithms}
W.~E. Hoover, ``{Algorithms for Confidence Circles and Ellipses},'' \emph{NOAA
  Technical Report}, 1984.

\bibitem{makhdoumi2017convergence}
A.~Makhdoumi and A.~Ozdaglar, ``{Convergence Rate of Distributed ADMM over
  Networks},'' \emph{IEEE Transactions on Automatic Control}, vol.~62, no.~10,
  pp. 5082--5095, 2017.

\bibitem{more1994line}
J.~J. Mor{\'e} and D.~J. Thuente, ``{Line Search Algorithms with Guaranteed
  Sufficient Decrease},'' \emph{ACM Transactions on Mathematical Software
  (TOMS)}, vol.~20, no.~3, pp. 286--307, 1994.

\bibitem{pan1999complexity}
V.~Y. Pan and Z.~Q. Chen, ``{The Complexity of the Matrix Eigenproblem},'' in
  \emph{Proceedings of the thirty-first annual ACM symposium on Theory of
  computing}, 1999, pp. 507--516.

\bibitem{stump2008connectivity}
E.~Stump, A.~Jadbabaie, and V.~Kumar, ``{Connectivity Management in Mobile
  Robot Teams},'' in \emph{2008 IEEE International Conference on Robotics and
  Automation}.\hskip 1em plus 0.5em minus 0.4em\relax IEEE, 2008, pp.
  1525--1530.

\bibitem{shah2018airsim}
S.~Shah, D.~Dey, C.~Lovett, and A.~Kapoor, ``{AirSim: High-fidelity Visual and
  Physical Simulation for Autonomous Vehicles},'' in \emph{Field and service
  robotics}.\hskip 1em plus 0.5em minus 0.4em\relax Springer, 2018, pp.
  621--635.

\bibitem{goretkin2013optimal}
G.~Goretkin, A.~Perez, R.~Platt, and G.~Konidaris, ``{Optimal Sampling-based
  Planning for Linear-quadratic Kinodynamic Systems},'' in \emph{2013 IEEE
  International Conference on Robotics and Automation}.\hskip 1em plus 0.5em
  minus 0.4em\relax IEEE, 2013, pp. 2429--2436.

\end{thebibliography}
